\pdfoutput=1
\documentclass[11pt]{article}

\usepackage[final]{acl}
\usepackage{times}
\usepackage{latexsym}
\usepackage{booktabs}
\usepackage{graphicx}
\usepackage{adjustbox}
\usepackage{amsmath}
\usepackage{url}
\usepackage{microtype}
\usepackage{multirow}
\usepackage{array}
\usepackage{flushend} %

\hypersetup{
  pdftitle={SAFE-KV: Causal Suffix Repair for Behavior-Preserving Non-Prefix KV Cache Reuse in RAG},
  pdfauthor={Xingjian Zhang},
}

\title{SAFE-KV: Causal Suffix Repair for Behavior-Preserving Non-Prefix KV Cache Reuse in RAG}
\author{Xingjian Zhang \\
  Carnegie Mellon University, Pittsburgh, PA, USA \\
  \texttt{xingjiaz@andrew.cmu.edu}}

\newcommand{\inventoryResultFiles}{648}
\newcommand{\inventoryAnswerFiles}{69}
\newcommand{\cpuTestsPassed}{135}
\newcommand{\cpuTestsSkipped}{1}

\begin{document}
\maketitle

\begin{abstract}
Retrieval-augmented generation (RAG) reuses the same documents across many requests, making non-prefix KV-cache reuse attractive. Yet a cached document state is not reusable text: it is a context-conditioned causal state whose keys and values depend on the prefix, position, and preceding documents under which it was computed. We formulate non-prefix KV reuse as a behavior-preservation problem with full prefill on the exact prompt as the reference, and introduce SAFE-KV, which repairs a composed cache by re-encoding the causal suffix from a selected start document. On Qwen2.5-7B-Instruct, prefix-conditioned composition preserves only 47.2\% of HotpotQA-500 and 37.8\% of 2Wiki-500 full-prefill answers; SAFE-KV with risk-guided K2 selection restores 99.8\%/98.4\% (donor-copy protocol) and 100.0\%/98.4\% with actual suffix forwarding. Document-only and fixed-window repairs recover far less, because the document states after the repaired region keep histories that omit it. Reuse without repair is exact up to four documents and degrades at sixteen, where suffix repair recovers 98--100\%. Faithful repair re-encodes most of the context, so in a single-request L40S study SAFE-KV's operating rule serves every workload by full prefill. SAFE-KV thus establishes the causal-state contract for non-prefix reuse and a validated repair mechanism that meets it.
\end{abstract}

\section{Introduction}

Retrieval-augmented generation (RAG) systems reread the same documents across many requests. Exact-prefix KV caching avoids recomputation only when the cached tokens form a prefix of the new prompt, so serving systems increasingly encode a document once and reuse its KV state wherever the document later appears. We argue that this is a question about model behavior, not tensor layout: a cached document state is a context-conditioned causal state, and reusing it outside the context that produced it changes what the model computes.

\paragraph{Problem in one paragraph.}
Let a rendered RAG prompt be $x=(p,d_1,\ldots,d_m,q)$: a static instruction prefix $p$, retrieved documents $d_1,\ldots,d_m$ in prompt order, and the query $q$. Full prefill runs the model over all of $x$ and yields a cache $C_{\mathrm{full}}$ in which every key and value was computed under the same causal history. \emph{Prefix-conditioned composition} instead encodes each $d_i$ under $p$ alone, at its target position, and concatenates the slices into a composed cache $C_{\mathrm{comp}}$. The two caches hold the same tokens at the same positions, but for $i>1$ the slice of $d_i$ in $C_{\mathrm{comp}}$ was not computed under $d_1,\ldots,d_{i-1}$: its context history is mismatched, and decoding $q$ from $C_{\mathrm{comp}}$ can give a different answer. SAFE-KV \emph{repairs} $C_{\mathrm{comp}}$ by choosing a start document $s$ and re-encoding the suffix $\mathrm{Suffix}(s)=\{d_i: i\geq s\}$ in order after the retained composed prefix; the query is always re-encoded. Section~\ref{sec:problem} formalizes the target (does the repaired cache reproduce the full-prefill answer?) and the cost (the fraction of context tokens re-encoded).

\begin{figure*}[t]
\centering
\includegraphics[width=0.92\linewidth]{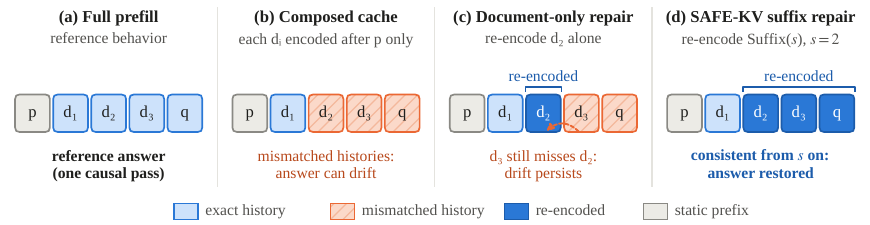}
\caption{SAFE-KV treats a non-prefix cache hit as a causal-state contract. (a)~Full prefill encodes every document under its true history. (b)~Prefix-conditioned composition encodes each document after the prefix alone, so later slices and the query miss their predecessors and the answer can drift. (c)~Re-encoding only the selected document leaves every later slice with a history that omits it. (d)~SAFE-KV re-encodes the causally closed suffix from the selected start $s$, restoring a consistent history from $s$ onward.}
\label{fig:setup}
\end{figure*}

The paper develops this view in three steps. \textbf{(1) Measuring the mismatch.} Causal dependence on earlier context is a basic property of transformers; we quantify its answer-level consequence for non-prefix document-cache composition on natural multi-hop RAG prompts, with full prefill on the exact prompt as the behavioral reference. The reference defines behavior, not truth: if the model is wrong under full prefill, a cache method that changes the answer has still changed the model, so gold EM/F1 are auxiliary. \textbf{(2) Causal suffix repair.} Every state after a mismatched document was computed under a history that omits it, so a faithful repair must be closed under causal dependence: re-encoding the start document alone, or a fixed window after it, leaves later states inconsistent by construction (Figure~\ref{fig:setup}c). SAFE-KV therefore re-encodes the suffix. \textbf{(3) Start selection.} Two training-free selectors, a query-overlap score and a risk score, and their union choose the start $s$, yielding operating points on a fidelity--cost trade-off.

\paragraph{Findings.}
On Qwen2.5-7B-Instruct, prefix-conditioned composition preserves only 47.2\% of HotpotQA-500 and 37.8\% of 2Wiki-500 full-prefill answers, and suffix repair with risk-guided K2 selection restores 99.8\%/98.4\% (donor copy) and 100.0\%/98.4\% (actual forwarding). Document-only repair and fixed windows recover much less: when the answer document comes first, re-encoding only that document is no better than no repair. Reuse is exact for up to four documents and degrades at sixteen, where repair recovers 96--100\%; without the construction's retrieval cues, query and risk K2 still preserve 91.0--96.0\%. Faithful repair re-encodes most of the context (94.8\%/92.9\% at K2), so in a single-request L40S study SAFE-KV's operating rule serves every workload by full prefill, and the measurements identify a later but still safe start as the lever for faster faithful reuse.

\paragraph{Contributions.}
\begin{enumerate}
    \item A causal-state formulation of non-prefix RAG KV reuse: full-prefill answer equivalence as the contract, explicit context-mismatch accounting, and an evaluation protocol that separates donor-copy diagnostics from deployable actual forwarding.
    \item Causal suffix repair, grounded in a causal-closure argument, with 7B-scale evidence that it restores full-prefill behavior where document-only and fixed-window repair do not.
    \item SAFE-KV operating points with executable selector definitions, and staleness $\times$ document-count, score-neutral, localized-baseline, and latency analyses that map where and at what cost faithful reuse is achieved.
    \item An open artifact with code, data, per-example outputs, and validators for every run.
\end{enumerate}

\section{Related Work}

\paragraph{Exact-prefix reuse and serving systems.}
Exact-prefix reuse is safe because the cache was produced under the same causal history; serving systems such as vLLM/PagedAttention, SGLang, and LMCache manage how that state is paged, scheduled, stored, and moved \citep{kwon2023efficientmemorymanagementlarge,zheng2024sglang,liu2025lmcacheefficientkvcache}. SAFE-KV asks what must hold semantically once reuse leaves the exact-prefix regime.

\paragraph{Non-prefix and modular KV reuse.}
Prompt Cache reuses attention states for recurring prompt modules \citep{gim2024promptcachemodularattention}; TurboRAG precomputes document caches for chunked RAG \citep{lu2024turborag}; KVLink adjusts positions and trains link tokens so independently encoded documents can be concatenated \citep{yang2025kvlinkacceleratinglargelanguage}. Two systems recompute selectively rather than reuse everything: CacheBlend fuses cached chunks and recomputes a subset of tokens with large KV deviation \citep{yao2025cacheblendfastlargelanguage}, and Cache-Craft manages chunk caches and recomputes selected tokens to maintain output quality when chunks appear in new contexts \citep{agarwal2025cachecraft}. These systems optimize serving throughput and report task accuracy. SAFE-KV differs in the evaluation target and the unit of repair: answer equivalence to full prefill on the exact prompt, a comparison of document-only, fixed-window, and suffix re-encoding, and matched random controls. Section~\ref{sec:results-local} compares suffix repair with an operational token-local baseline built in the same spirit.

\paragraph{KV compression, long context, and RAG robustness.}
KV compression and long-context methods start from a valid causal state and ask how to store, compress, or extend it \citep{liu2024kivituningfreeasymmetric2bit,zhang2023h2o,xiao2023streamingllm,li2024snapkv,su2021roformer,peng2023yarn}; SAFE-KV asks the prior question of whether a composed state is behaviorally equivalent to full prefill. Retrieval, multi-hop, and robustness benchmarks \citep{lewis2021retrievalaugmentedgenerationknowledgeintensivenlp,yang2018hotpotqadatasetdiverseexplainable,ho2020constructingmultihopqadataset,liu2023lostmiddle,chen2023rgb,cuconasu2024powernoise} motivate our data; we use them to test cache-state preservation rather than QA accuracy. Appendix~\ref{sec:app-reference-map} extends this map; Appendix~\ref{sec:app-related-comparison} tabulates the differences.

\section{Problem Setup}
\label{sec:problem}

\paragraph{Full-prefill and composed states.}
Let a RAG prompt be the sequence $x=(p,d_1,\ldots,d_m,q)$, where $p$ is the static chat or instruction prefix, $d_1,\ldots,d_m$ are retrieved document blocks in prompt order, and $q$ is the query suffix. Full prefill runs the transformer over the entire rendered prompt and produces a cache $C_{\mathrm{full}}$ in which every key and value was computed under the same causal history.

A non-prefix cache reuse method builds a different object. Prefix-conditioned composition encodes each document under the static prefix, at the global position it occupies in the target prompt, stores the document slice, and concatenates the slices into a composed cache $C_{\mathrm{comp}}$. Token order and position IDs match the rendered prompt, but for $i>1$ the slice of $d_i$ was not produced under the preceding documents $d_1,\ldots,d_{i-1}$. We call this a \emph{context mismatch}; ``stale'' below always refers to this provenance mismatch, never to document age. The question is whether decoding from $C_{\mathrm{comp}}$ preserves the behavior of decoding after full prefill.

\paragraph{Behavioral reference and metric.}
SAFE-KV uses full prefill as the behavioral reference. Both caches are decoded greedily with at most 16 new tokens and the same stop strings; the primary metric is
\[
\mathbf{1}\{\mathrm{norm}(a_{\mathrm{full}})=\mathrm{norm}(a_{\mathrm{method}})\},
\]
where $\mathrm{norm}$ lower-cases, replaces ASCII punctuation by spaces, removes the articles \emph{a/an/the}, and collapses whitespace. We call this \emph{answer preservation} (or answer equivalence); it is a normalized short-answer exact match, not equality of raw strings, token sequences, or hidden states. Where a run also records raw-string and generated-token-sequence equality we report them. This target is distinct from task correctness: if the model is gold-wrong under full prefill, answer preservation still measures whether cache reuse changed the computation. We report gold EM/F1 as auxiliary metrics, \emph{changed-wrong} counts (answers that differ from full prefill and also fail gold exact match), and first-answer-token KL as a distributional diagnostic.

\paragraph{Repair from a selected start.}
Repair starts from $C_{\mathrm{comp}}$ and re-encodes a region of the document context. Given a selected set of document ids $S$, let $s=\min\{i: d_i\in S\}$ be the earliest selected prompt position; SAFE-KV re-encodes the suffix
\[
\mathrm{Suffix}(s)=\{d_i : i\geq s\}
\]
in order, after the retained composed prefix $(p,d_1,\ldots,d_{s-1})$, and then encodes the query. We refer to $s$ as the \emph{selected suffix-recompute start}. Document-only repair re-encodes only $d_s$; a downstream window of size $w$ re-encodes $d_s,\ldots,d_{\min(s+w-1,m)}$.

\paragraph{Causal closure.}
In a causal decoder the key and value at position $t$ depend only on the tokens at positions $\leq t$. Every slice $d_j$ with $j>s$ in $C_{\mathrm{comp}}$ was computed under a history that omits $d_s$, so it stays mismatched however well $d_s$ itself is repaired. Call a set of repaired documents \emph{causally closed} if it contains every document after each of its members: $\mathrm{Suffix}(s)$ is the smallest causally closed set containing $d_s$, whereas a single document or a fixed window leaves later slices mismatched by construction. Because the retained prefix is itself composed, suffix repair coincides with full prefill only when $s=1$; how late $s$ may start while preserving the answer is an empirical, per-example question for the selectors (Section~\ref{sec:selectors}).

\paragraph{Two execution protocols.}
Repair can be executed in two ways, and every table states which one it uses. Under \emph{actual forwarding}, the deployable path, the selected contiguous suffix is run through the model from the retained composed prefix cache and the resulting key/value states replace the composed slices. Under \emph{donor copy}, an offline diagnostic, the corresponding slices are copied from a separately computed $C_{\mathrm{full}}$. Donor copy inserts states computed under the exact full history, whereas actual forwarding computes them under the retained composed prefix. Tables~\ref{tab:external-main}, \ref{tab:local-vs-suffix}, and~\ref{tab:position} use donor copy; Tables~\ref{tab:actual-forwarding}, \ref{tab:local-operational}, \ref{tab:staleness}, and~\ref{tab:latency-main} use actual forwarding; Appendix~\ref{sec:app-external} reconciles the two on the same run.

\paragraph{Budget.}
We report the \emph{recomputed context ratio}: re-encoded document-block tokens divided by the static-prefix-plus-document context length (the query is excluded from both). The ratio measures work rather than time, so Section~\ref{sec:results-latency} measures time to first logit directly.

\section{SAFE-KV Causal Suffix Repair}

\subsection{Prefix-conditioned composition}

SAFE-KV starts from the strongest simple non-prefix baseline, prefix-conditioned document composition. Each document $d_i$ is encoded under the same static instruction/chat prefix $p$ used by the full prompt and at its target global position, and only the document cache slice is stored. At reuse time the static prefix cache and document slices are assembled in the target order, and the query is evaluated against the composed context cache. This baseline removes prefix and position mismatch and isolates the missing cross-document history.

\subsection{Selecting the suffix-recompute start}
\label{sec:selectors}

SAFE-KV uses two lightweight, training-free selectors to choose the start $s$. Both are defined here exactly as implemented (\texttt{safekv/doc\_selector.py}); Appendix~\ref{sec:app-selector} lists the regular expressions, the stopword list, and a feature-by-feature map to the code. Let $T(\cdot)$ be the set of lower-cased alphanumeric tokens of a string after removing a fixed list of 21 stopwords, and let $\mathrm{Ent}(\cdot)$ be the set of lower-cased entity strings extracted by regular expressions (multi-word capitalized spans, all-capital tokens, number/time-like tokens, and single capitalized words). Write $t_i$ for the title of $d_i$ concatenated with its text.

\paragraph{Query selector.}
The query score is the recall of unique question terms in the document,
\begin{equation}
Q_i=\frac{|T(q)\cap T(t_i)|}{|T(q)|},
\label{eq:query}
\end{equation}
with $Q_i=0$ if $T(q)$ is empty.

\paragraph{Risk selector.}
The risk score is a fixed weighted sum of five features,
\begin{equation}
\begin{aligned}
R_i={}&0.25\,\hat r_i+0.30\,Q_i+0.20\,E_i\\&+0.15\,I^{\mathrm{conf}}_i+0.10\,\hat X_i,
\end{aligned}
\label{eq:risk}
\end{equation}
where $\hat r_i$ is the document's retrieval score min--max normalized within the example (all documents receive $1$ if the scores are equal and positive, $0$ if they are all zero); $E_i=|\mathrm{Ent}(q)\cap\mathrm{Ent}(t_i)|/|\mathrm{Ent}(q)|$ is question-entity recall; $I^{\mathrm{conf}}_i\in\{0,1\}$ marks conflict candidates, documents that contain a number- or date-like answer candidate and either a comparison cue in the question (\emph{according to, official, latest, current, schedule, trusted}) or $Q_i\geq 0.25$, provided that the candidate value sets of at least two such documents disagree; and $\hat X_i$ is the number of entities of $d_i$ that also occur in at least one other document, min--max normalized within the example. The weights are fixed in the implementation and shared by every reported run (no hyperparameter search); Appendix~\ref{sec:app-selector} maps each term to the code.

\paragraph{Selection, hybrid, and start.}
Selector $K$ returns the $K$ documents with the largest score, ties broken by ascending document id. Document ids equal prompt positions in the external conversions, their score-neutral variants, and the nested staleness inputs, so a tie there selects the earlier document (a longer suffix); in the synthetic diagnostics evidence documents carry the smallest ids, so a tie there favors the evidence document. Ties are rare for the risk score and frequent for the query score, whose high-overlap distractors share the question terms; Appendix~\ref{sec:app-selector} quantifies this for every dataset, and Sections~\ref{sec:results-main}, \ref{sec:results-position}, and~\ref{sec:results-stale} account for it. The hybrid selector takes the union of the query top-1 and the risk top-1. In all cases the suffix-recompute start is $s=\min\{i: d_i\in S\}$, so hybrid K1 repairs from the earlier of two plausible starts and coincides with K1 when the selectors agree.

\paragraph{Inputs and scope.}
The selectors read only the question text, the document titles and texts, the document order, and the per-example retrieval scores; they never read gold evidence labels, gold answers, the full-prefill cache, logits, or answer, or any sampled mismatch mask. On the original external conversion the retrieval score is construction-assigned (1.0 for supporting documents, at most 0.55 for distractors) and therefore correlates with the gold labels, so Section~\ref{sec:results-main} also evaluates score-neutral shuffled inputs in which every retrieval score is zero. The selectors are operating policies for choosing $s$, to be validated per workload under the operating rule; we keep them training-free because a synthetic-trained learned selector did not transfer to the external data (Appendix~\ref{sec:app-selector}).

\subsection{Causal suffix repair and operating rule}
\label{sec:operating-rule}

Given $S$, SAFE-KV builds the prefix-conditioned composed cache, sets $s$ to the earliest selected position, runs the contiguous suffix from $s$ through the model under the retained prefix (actual forwarding) or, in the offline diagnostic, copies the corresponding full-prefill slices (donor copy), and decodes the query greedily (Section~\ref{sec:problem}). The main operating points are risk-guided suffix K1 (lowest budget), hybrid query+risk suffix K1 (balanced), and risk-guided suffix K2 (highest fidelity).

\paragraph{Operating rule.}
SAFE-KV serves a request from a repaired composed cache only when a validated operating point for that workload both preserves answers at the declared target and reduces measured time to first logit; otherwise the request is served by full prefill. The rule makes behavior preservation a hard constraint and speed an opportunistic gain; it is applied to measurements rather than a pre-registered threshold, and on every workload of Section~\ref{sec:results-latency} it selects full prefill.

\subsection{Baselines and ablation policies}

We compare suffix repair against document-only repair (only the selected documents), downstream windows (the selected document plus the next $w-1$), random suffix controls (random documents with the same $K$, then suffix repair), oracle repairs that use gold evidence documents to diagnose the mechanism, and, in Section~\ref{sec:results-local}, an operational token-local baseline that ranks context tokens by the query's attention over the stale cache. Table~\ref{tab:design-ledger} in Appendix~\ref{sec:app-design-ledger} records the broader development ledger.

\section{Experiments}

\paragraph{Main external evaluation and its construction.}
The main experiments use Qwen2.5-7B-Instruct \citep{qwen2025qwen25technicalreport} on two multi-hop RAG conversions, HotpotQA-500 and 2Wiki-500 \citep{yang2018hotpotqadatasetdiverseexplainable,ho2020constructingmultihopqadataset}. We remove yes/no questions, map each context paragraph to a document, and keep distractor paragraphs. The conversion is \emph{gold-informed}: supporting paragraphs are placed first in the prompt and receive retrieval score $1.0$, while distractors follow in order of lexical overlap with the question and receive scores of at most $0.55$ (Appendix~\ref{sec:app-data}). This controlled setting isolates repair fidelity. Because the risk selector reads the retrieval score, we test selection without these cues on \emph{score-neutral shuffled} variants of the same examples, in which every retrieval score is zero and documents are shuffled with per-example order seeds 101, 202, and 303 (100 examples each).

\paragraph{Controlled diagnostics.}
Synthetic diagnostics isolate failure modes with single-evidence, multi-hop, conflict, boundary-sensitive, and distractor-heavy examples; noisy variants add high-overlap distractors while preserving evidence labels, and evidence documents also receive retrieval score $1.0$ in these files. The staleness-by-document-count benchmark nests single-evidence examples into $N\in\{1,2,4,8,16\}$ documents with a label-independent layout (seed 101) and zero retrieval scores. The operational localized-repair baseline uses separate HotpotQA conversions with ten paragraphs per example in natural, shuffled, and BM25 order, built without gold labels for ranking.

\paragraph{Models and hardware.}
The main model is Qwen2.5-7B-Instruct (FP16); Mistral-7B-Instruct-v0.3 \citep{jiang2023mistral} is evaluated on a balanced noisy diagnostic set. Small-model development diagnostics appear in the appendix. The donor-copy results of Tables~\ref{tab:external-main}, \ref{tab:local-vs-suffix}, and~\ref{tab:position}, their appendix analyses, and the development diagnostics used an NVIDIA L40S-class GPU with PyTorch 2.4.1+cu124; all actual-forwarding, reconciliation, staleness, latency, and oracle-diagnostic runs used one NVIDIA L40S with PyTorch 2.8.0+cu128; the token-local baseline used one NVIDIA RTX PRO 6000 Blackwell GPU with PyTorch 2.8.0+cu128; Transformers 4.57.6 throughout.

\paragraph{Methods, protocols, and metrics.}
We compare full prefill, prefix-conditioned composition, query suffix K1/K2, risk suffix K1/K2, hybrid suffix K1 (the main SAFE-KV operating points are risk K1, hybrid K1, and risk K2), random suffix controls, document-only patching, downstream windows, oracle repair variants, and the query-attention token-local baseline. Every table states whether it uses donor copy or actual forwarding (Section~\ref{sec:problem}). The primary metric is normalized answer preservation relative to full prefill under greedy decoding with 16 maximum new tokens, the \texttt{chat\_rag} prompt format, and manual position IDs. We also report gold EM/F1, changed-wrong counts, the recomputed context ratio, first-token KL, generated-token exactness where recorded, paired bootstrap intervals, held-out calibration, and, for latency, time to first logit with selector planning included.

\section{Results}

\subsection{Main 7B external results}
\label{sec:results-main}

Table~\ref{tab:external-main} reports the main result (donor-copy protocol, gold-evidence-first construction). Prefix-conditioned composition is not behavior preserving on natural multi-hop RAG prompts: it preserves 47.2\% of full-prefill answers on HotpotQA-500 and 37.8\% on 2Wiki-500, with 241 and 272 changed-wrong answers. Suffix repair closes the gap: risk K1, hybrid K1, and risk K2 reach 99.6\%, 99.6\%, and 99.8\% on HotpotQA and 97.0\%, 97.4\%, and 98.4\% on 2Wiki; random suffix K2 is strong but clearly weaker (91.8\%/83.4\%), and risk K2 matches full-prefill gold EM (54.4 on HotpotQA, 45.2 on 2Wiki). The ascending-id tie-break decides the start for risk K1 in 7 and 6 of 500 examples (HotpotQA/2Wiki) and for risk K2 in 1 and 0, but for hybrid K1 in 38 and 50 and for query K2 in 183 and 174 (Appendix~\ref{sec:app-selector}), so selector conclusions rest on the risk rows and the score-neutral runs below.

\begin{table*}[t]
\centering
\footnotesize
\begin{adjustbox}{max width=\textwidth}
\begin{tabular}{llrrrrr}
\toprule
Data & Method & Ans. eq. & Gold EM & Gold F1 & Changed-wrong & Patch \\
\midrule
HotpotQA-500 & Prefix-comp. & 0.472 & 0.354 & 0.489 & 241 & 0.000 \\
HotpotQA-500 & Query suffix K2 & 0.958 & 0.534 & 0.704 & 18 & 0.883 \\
HotpotQA-500 & Random suffix K2 & 0.918 & 0.528 & 0.696 & 35 & 0.814 \\
HotpotQA-500 & Risk suffix K1 & 0.996 & 0.542 & 0.708 & 2 & 0.892 \\
HotpotQA-500 & Hybrid suffix K1 & 0.996 & 0.542 & 0.708 & 2 & 0.904 \\
HotpotQA-500 & Risk suffix K2 & 0.998 & 0.544 & 0.709 & 1 & 0.948 \\
\midrule
2Wiki-500 & Prefix-comp. & 0.378 & 0.312 & 0.386 & 272 & 0.000 \\
2Wiki-500 & Query suffix K2 & 0.972 & 0.444 & 0.538 & 12 & 0.885 \\
2Wiki-500 & Random suffix K2 & 0.834 & 0.434 & 0.517 & 66 & 0.711 \\
2Wiki-500 & Risk suffix K1 & 0.970 & 0.442 & 0.536 & 12 & 0.813 \\
2Wiki-500 & Hybrid suffix K1 & 0.974 & 0.442 & 0.536 & 11 & 0.842 \\
2Wiki-500 & Risk suffix K2 & 0.984 & 0.452 & 0.546 & 6 & 0.929 \\
\bottomrule
\end{tabular}
\end{adjustbox}
\caption{Main 7B external results with Qwen2.5-7B-Instruct: donor-copy protocol, gold-evidence-first construction (Appendix~\ref{sec:app-data}). Ans.\ eq.\ is normalized answer preservation relative to full prefill; Changed-wrong counts answers that differ from full prefill and fail gold exact match; Patch is the recomputed context ratio. Actual-forwarding values for the same inputs: Table~\ref{tab:actual-forwarding}(a).}
\label{tab:external-main}
\end{table*}

\begin{table}[t]
\centering
\footnotesize
\setlength{\tabcolsep}{3pt}
\begin{adjustbox}{max width=\linewidth}
\begin{tabular}{lrrrrrr}
\toprule
 & \multicolumn{3}{c}{HotpotQA} & \multicolumn{3}{c}{2Wiki} \\
\cmidrule(lr){2-4}\cmidrule(lr){5-7}
Method & Eq. & Tok. & Rec. & Eq. & Tok. & Rec. \\
\midrule
\multicolumn{7}{l}{\emph{(a) Gold-evidence-first construction, $N=500$}} \\
Risk K1 & 0.998 & 0.998 & 0.892 & 0.966 & 0.958 & 0.813 \\
Risk K2 & 1.000 & 1.000 & 0.948 & 0.984 & 0.980 & 0.929 \\
\midrule
\multicolumn{7}{l}{\emph{(b) Score-neutral shuffled order, 3 seeds $\times$ 100 (means)}} \\
Risk K1 & 0.830 & 0.820 & 0.564 & 0.773 & 0.757 & 0.622 \\
Hybrid K1 & 0.910 & 0.903 & 0.724 & 0.843 & 0.830 & 0.714 \\
Query K2 & 0.960 & 0.957 & 0.865 & 0.953 & 0.947 & 0.863 \\
Risk K2 & 0.917 & 0.913 & 0.762 & 0.910 & 0.897 & 0.801 \\
\bottomrule
\end{tabular}
\end{adjustbox}
\caption{Actual suffix forwarding with Qwen2.5-7B-Instruct (the selected contiguous suffix is re-run through the model from the retained composed prefix; no donor-cache copying). All rows are suffix repair (``Risk K1'' = risk suffix K1, etc.). Eq.\ is normalized answer preservation relative to full prefill; Tok.\ is generated-token-sequence equality; Rec.\ is the mean fraction of context tokens re-forwarded. Panel (a) uses the same gold-evidence-first inputs as Table~\ref{tab:external-main}, so the two tables differ only in protocol (changed-wrong counts: 1/0 on HotpotQA, 12/5 on 2Wiki for K1/K2). Panel (b) removes the construction cues: every retrieval score is zero and documents are shuffled with order seeds 101/202/303 (100 examples each); values are means over seeds, per-seed values in Table~\ref{tab:app-neutral-seed}. Random suffix and query suffix K1 were not run under shuffled order. One NVIDIA L40S, PyTorch~2.8.0.}
\label{tab:actual-forwarding}
\end{table}

Table~\ref{tab:actual-forwarding} evaluates the deployable protocol. Panel (a) re-runs the risk operating points on the same inputs with actual suffix forwarding: 99.8\%/100.0\% on HotpotQA and 96.6\%/98.4\% on 2Wiki for K1/K2, at 89.2\%/94.8\% and 81.3\%/92.9\% recomputation. The two protocols agree closely (a same-run reconciliation finds identical answers on HotpotQA and differences on 17 and 7 of 500 2Wiki outputs; Appendix~\ref{sec:app-external}), which confirms suffix repair as a fidelity mechanism; at these high-fidelity points it re-encodes most of the context. Panel (b) removes the construction cues: under score-neutral shuffled order, query K2 preserves 96.0\%/95.3\% at 86.5\%/86.3\% recomputation and risk K2 91.7\%/91.0\% at 76.2\%/80.1\%, while risk K1 falls to 83.0\%/77.3\% and hybrid K1 to 91.0\%/84.3\%. The mechanism is identical in both panels; what changes is whether the selected start precedes the evidence.

\subsection{Local repair versus suffix repair}
\label{sec:results-local}

\begin{figure}[t]
\centering
\includegraphics[width=0.86\linewidth]{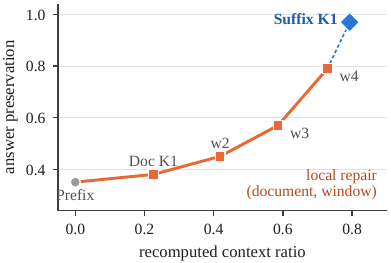}
\caption{Local repair versus suffix repair on 2Wiki-100 (values of Table~\ref{tab:local-vs-suffix}; donor-copy protocol; risk selector fixed). Document-only (Doc K1) and window (w2--w4) re-encoding recover answer preservation gradually; the suffix from the same start (Suffix K1) recovers 97.0\% at a budget close to w4's. Plotted values: \texttt{figure\allowbreak\_local\allowbreak\_vs\allowbreak\_suffix.data.csv} (artifact).}
\label{fig:local-vs-suffix}
\end{figure}

Table~\ref{tab:local-vs-suffix} compares repair units on 2Wiki-100 with the risk selector fixed. Replacing only the selected document barely helps (35.0\% to 38.0\%); a four-document window reaches 79.0\% at a 72.9\% budget, whereas suffix repair reaches 97.0\% at 79.3\% and cuts changed-wrong answers from 57 to 2 (Figure~\ref{fig:local-vs-suffix}). The high-noise synthetic set shows the same ordering: prefix composition preserves 41.6\%, oracle evidence-document patching 54.4\%, random K4 document patching 58.0\%, oracle suffix repair 93.6\%, and all-document patching 99.6\% (Appendix~\ref{sec:app-closure}). Each additional downstream document raises preservation (38.0\%, 45.0\%, 57.0\%, and 79.0\% for the document and windows w2--w4), and only the causally closed suffix closes the gap: re-encoding the downstream context, not the evidence document, is what recovers the answer, exactly as causal closure predicts.

\begin{table}[t]
\centering
\footnotesize
\setlength{\tabcolsep}{3.5pt}
\begin{adjustbox}{max width=\linewidth}
\begin{tabular}{lrrrr}
\toprule
Method & Ans. eq. & Chg.-wrong & Patch & KL \\
\midrule
Prefix-comp. & 0.350 & 57 & 0.000 & 4.974 \\
Risk doc K1 & 0.380 & 54 & 0.226 & 4.336 \\
Risk downstream w2 & 0.450 & 45 & 0.419 & 3.604 \\
Risk downstream w3 & 0.570 & 37 & 0.586 & 2.659 \\
Risk downstream w4 & 0.790 & 18 & 0.729 & 1.082 \\
Risk suffix K1 & 0.970 & 2 & 0.793 & 0.299 \\
\bottomrule
\end{tabular}
\end{adjustbox}
\caption{Local repair versus suffix repair on 2Wiki-100 (first 100 examples; donor-copy protocol; gold-evidence-first construction; risk selector fixed). Fixed downstream windows improve gradually but remain far below suffix repair; w4 spends nearly suffix-level budget while leaving many more changed-wrong answers.}
\label{tab:local-vs-suffix}
\end{table}

\begin{table}[t]
\centering
\footnotesize
\setlength{\tabcolsep}{4pt}
\begin{adjustbox}{max width=\linewidth}
\begin{tabular}{lrrr}
\toprule
 & Prefix- & Random & Query-attn. \\
Order & comp. & local & local \\
\midrule
natural & 0.330 & 0.590 & 0.850 \\
shuffled & 0.430 & 0.607 & 0.790 \\
BM25 & 0.300 & 0.537 & 0.840 \\
\midrule
pooled (300) & 0.353 & 0.578 & 0.827 \\
\bottomrule
\end{tabular}
\end{adjustbox}
\caption{Operational token-local repair on HotpotQA (100 examples per order; natural, shuffled, and BM25 document orders built without gold labels), Qwen2.5-7B-Instruct, actual forwarding of the selected token spans from the stale composed cache. Query-attn.\ local ranks context tokens by the query's attention over the stale cache and recomputes the top spans within a 60\% document-token budget; Random local recomputes equal-budget random spans (mean over three seeds). Values are normalized answer preservation relative to the all-fresh reference; the recomputed fraction of context tokens is 0.589 in every row. Runs: one NVIDIA RTX PRO 6000 Blackwell GPU, PyTorch~2.8.0. Full budget curves: Table~\ref{tab:app-local-full}.}
\label{tab:local-operational}
\end{table}

We also built an operational token-local baseline in the spirit of CacheBlend and Cache-Craft \citep{yao2025cacheblendfastlargelanguage,agarwal2025cachecraft}: the query is run against the stale composed cache, context tokens are ranked by the query's attention over that cache, and the top spans are re-encoded within a document-token budget (Table~\ref{tab:local-operational}; Appendix~\ref{sec:app-algorithms}). Selection uses no gold label, full-prefill state, or mismatch information. Pooled over 300 HotpotQA examples in natural, shuffled, and BM25 order, it preserves 82.7\% of full-prefill answers at a 60\% budget, against 57.8\% for equal-budget random spans and 35.3\% without repair. Informed token placement thus helps substantially, but it still leaves more drift than the high-fidelity suffix operating points, consistent with causal closure: unselected spans downstream of a mismatch keep their stale histories. Our localized-repair conclusions concern this baseline and the units of Table~\ref{tab:local-vs-suffix}; we did not run the CacheBlend or Cache-Craft systems themselves.

\subsection{Position sensitivity}
\label{sec:results-position}

\begin{figure}[t]
\centering
\includegraphics[width=0.86\linewidth]{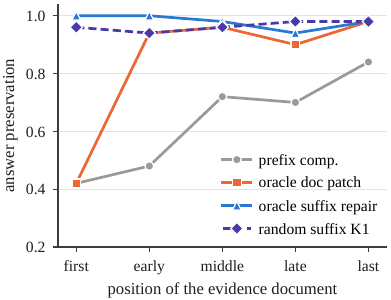}
\caption{Position sensitivity (values of Table~\ref{tab:position}; donor-copy protocol; 50 examples per position). With the evidence document first, patching only that document is no better than prefix composition, because every later slice keeps its mismatched history; suffix repair removes most of the position dependence. Plotted values: \texttt{figure\allowbreak\_position\allowbreak\_sweep.data.csv} (artifact).}
\label{fig:position-sweep}
\end{figure}

Table~\ref{tab:position} fixes the answer-bearing document and moves it through a high-noise prompt. Prefix composition preserves 42.0\% when the evidence is first, 48.0\% when early, and 84.0\% when last: in the first position every later slice was encoded without the evidence. Oracle document patching stays at 42.0\% in that case, the no-repair rate, while oracle suffix repair reaches 98.0\% overall and random suffix K1 96.4\%. This is causal closure in its cleanest form (Figure~\ref{fig:position-sweep}): the evidence state itself is correct, yet every downstream state still omits it. The ``query suffix K1'' row equals the oracle-suffix row because the ascending-id tie-break selects the evidence document (id 0) in all 250 cases (Appendix~\ref{sec:app-selector}).

\begin{table}[t]
\centering
\footnotesize
\setlength{\tabcolsep}{3pt}
\begin{adjustbox}{max width=\linewidth}
\begin{tabular}{lrrrrrr}
\toprule
Method & Overall & First & Early & Middle & Late & Last \\
\midrule
Prefix-comp. & 0.632 & 0.420 & 0.480 & 0.720 & 0.700 & 0.840 \\
Oracle doc patch & 0.840 & 0.420 & 0.940 & 0.960 & 0.900 & 0.980 \\
Oracle suffix & 0.980 & 1.000 & 1.000 & 0.980 & 0.940 & 0.980 \\
Query suffix K1 & 0.980 & 1.000 & 1.000 & 0.980 & 0.940 & 0.980 \\
Random suffix K1 & 0.964 & 0.960 & 0.940 & 0.960 & 0.980 & 0.980 \\
\bottomrule
\end{tabular}
\end{adjustbox}
\caption{Position sensitivity on high-noise synthetic examples (50 per position, 19 documents; donor-copy protocol). When the evidence document is first, patching only that document is no better than prefix composition; suffix repair removes most position dependence. ``Query suffix K1'' coincides with ``Oracle suffix'' because the id tie-break selects the evidence document in every example; on this sweep that row is effectively an oracle-start result (Section~\ref{sec:results-position}).}
\label{tab:position}
\end{table}

\subsection{Staleness probability and document count}
\label{sec:results-stale}

Table~\ref{tab:staleness} varies the number of documents $N\in\{1,2,4,8,16\}$ and the mismatch probability $p\in\{0,0.25,0.5,0.75,1\}$ on a controlled single-evidence benchmark (50 paired examples, three mask seeds, actual forwarding). Here $p$ is the probability that an eligible slice was encoded from the static prefix rather than its exact preceding history. Reuse without repair is exact in every cell with $N\leq 4$ and at least 99.3\% at $N=8$, so short contexts can be reused directly. At $N=16$, no-repair preservation across $p=0/0.25/0.5/0.75/1$ is 100.0/92.0/92.7/94.0/80.0\%, not monotone in $p$. In the hardest cell ($N{=}16$, $p{=}1$) a four-document window from the query start reaches 96.0\% at 24.0\% recomputation, risk K1 98.0\% at 68.6\%, and risk K2 100.0\% at 79.6\%, with paired gains over no repair of +16 [6, 26], +18 [6, 30], and +20 [10, 32] points. Because retrieval scores are zero and the nested documents tie on the lexical features, the query selector starts at the first document in 43 of 50 examples, while risk K1 and K2 start later in 31 and 21 examples; the query rows therefore compare early-start units, and the best unit depends on the budget.

\begin{table}[t]
\centering
\footnotesize
\setlength{\tabcolsep}{3pt}
\begin{adjustbox}{max width=\linewidth}
\begin{tabular}{lrrrrrr}
\toprule
 & \multicolumn{5}{c}{mismatch probability $p$} & \\
\cmidrule(lr){2-6}
 & 0 & 0.25 & 0.5 & 0.75 & 1 & Recomp. \\
\midrule
\multicolumn{7}{l}{\emph{No repair, by document count $N$}} \\
1 & 100.0 & 100.0 & 100.0 & 100.0 & 100.0 \\
2 & 100.0 & 100.0 & 100.0 & 100.0 & 100.0 \\
4 & 100.0 & 100.0 & 100.0 & 100.0 & 100.0 \\
8 & 100.0 & 99.3 & 99.3 & 99.3 & 100.0 \\
16 & 100.0 & 92.0 & 92.7 & 94.0 & 80.0 \\
\midrule
\multicolumn{7}{l}{\emph{Repair policies at $N=16$ (actual forwarding)}} \\
Query doc K1 & 100.0 & 92.0 & 92.7 & 96.0 & 84.0 & 5.5 \\
Query w4 K1 & 100.0 & 91.3 & 93.3 & 95.3 & 96.0 & 24.0 \\
Query suffix K1 & 100.0 & 100.0 & 100.0 & 100.0 & 100.0 & 89.8 \\
Risk suffix K1 & 100.0 & 96.7 & 98.0 & 98.7 & 98.0 & 68.6 \\
Risk suffix K2 & 100.0 & 98.0 & 99.3 & 100.0 & 100.0 & 79.6 \\
\bottomrule
\end{tabular}
\end{adjustbox}
\caption{Staleness probability $\times$ document count: normalized answer preservation (\%) relative to the all-fresh reference on a controlled single-evidence benchmark with $N$ documents (nested from the Noise16 pool, layout seed 101, retrieval scores zero), 50 paired examples, three mismatch-mask seeds. $p$ is the probability that an eligible slice (position $\geq 2$) was encoded from the static prefix instead of its exact preceding history; the query is always recomputed and no selector sees the mask. ``Query doc K1'' and ``Query w4 K1'' are the query doc-only and query window-w4 policies; Recomp.\ is the mean recomputed context fraction at $N{=}16$. Actual forwarding, Qwen2.5-7B-Instruct, one NVIDIA L40S; full grid with clustered intervals in Table~\ref{tab:app-stale-full}.}
\label{tab:staleness}
\end{table}

\subsection{Selector operating points}
\label{sec:results-selector}

\begin{table*}[t]
\centering
\footnotesize
\begin{adjustbox}{max width=\textwidth}
\begin{tabular}{lrrrrrrr}
\toprule
Workload (order) & Prompt tok. & Full (ms) & Risk K1 (ms) & Repair (\%) & Paired speedup [95\% CI] & Faster & p95 full / K1 (ms) \\
\midrule
HotpotQA, original order & 760 & 65.5 & 93.6 & 91.7 & 0.655 [0.624, 0.675] & 0\% & 88.7 / 129.6 \\
2Wiki, original order & 775 & 65.9 & 88.5 & 81.4 & 0.667 [0.647, 0.704] & 0\% & 118.9 / 147.1 \\
2Wiki, score-neutral shuffled & 775 & 65.6 & 78.1 & 58.6 & 0.722 [0.678, 0.811] & 10\% & 119.0 / 147.1 \\
$N{=}16$ nested, $p{=}1$ & 661 & 62.1 & 87.8 & 72.3 & 0.702 [0.692, 0.726] & 0\% & 63.0 / 90.9 \\
Noise64 (67 documents) & 2540 & 199.4 & 174.6 & 51.9 & 1.149 [1.027, 1.426] & 72\% & 219.4 / 260.6 \\
\bottomrule
\end{tabular}
\end{adjustbox}
\caption{Time to first logit (ms) with actual suffix forwarding, selector scoring and suffix-start planning included, on one NVIDIA L40S (Qwen2.5-7B-Instruct, FP16, SDPA, batch size 1, composed cache resident; cache construction, lookup, transfer, scheduling, token emission, and concurrency excluded). Fifty examples per workload, five timed repeats; medians over per-example medians, p95 over all 250 calls; paired speedup is full/Risk-K1 per example with an example bootstrap; ``Faster'' is the fraction of examples improved. Orders: original = inputs of Table~\ref{tab:external-main}; shuffled = score-neutral seed 101; $N{=}16$ and Noise64 synthetic. On Noise64, actual forwarding preserves only 44\% of answers on the same examples (Table~\ref{tab:app-noise64-gate}). All methods: Tables~\ref{tab:app-lat-full}--\ref{tab:app-plan}.}
\label{tab:latency-main}
\end{table*}

Table~\ref{tab:selector-budget} (Appendix~\ref{sec:app-selector}) lists the operating points on 2Wiki-500 (risk K1 97.0\% at 81.3\%, hybrid K1 97.4\% at 84.2\%, risk K2 98.4\% at 92.9\%) and on the Mistral-7B noisy diagnostic, where query K1 (90.8\%) beats risk K1 (85.2\%), hybrid K1 reaches 92.8\%, query and risk K2 both reach 99.6\%, and random suffix K2 reaches 100.0\%. Suffix repair thus transfers to Mistral at K2 whatever the selector, while the K1 ranking changes (the Mistral query and hybrid rows are largely decided by the id tie-break); with Table~\ref{tab:actual-forwarding}(b), the selectors are operating policies to be validated per workload, as the operating rule requires.

\subsection{Latency under actual forwarding}
\label{sec:results-latency}

Table~\ref{tab:latency-main} reports time to first logit on one L40S for five workloads with actual suffix forwarding and selector planning inside the timed region (all methods in Appendix~\ref{sec:app-cost}). On HotpotQA and 2Wiki in original order, on 2Wiki in shuffled order, and on the $N{=}16$ benchmark, risk K1 is slower than full prefill at the paired median (speed-ups 0.655--0.722; 0--10\% of examples faster) because it re-encodes 59--92\% of the context and then pays planning and query-forward costs; query K1, hybrid K1, and risk K2 behave the same way. Only the 67-document Noise64 workload shows a median crossover, from 199.4 to 174.6\,ms at 51.9\% recomputation (paired speed-up 1.149 [1.027, 1.426]); on the same 50 examples, however, actual forwarding preserves only 44\% [30\%, 58\%] of full-prefill answers (gold EM 44\% vs.\ 92\%), and p95 latency rises from 219.4 to 260.6\,ms. Under the operating rule of Section~\ref{sec:operating-rule}, every measured workload is therefore served by full prefill. Selector scoring is cheap (query 0.10--0.42\,ms, risk 0.9--3.3\,ms median per call); the cost is re-encoding. These are batch-1, resident-cache microbenchmarks without a serving stack or concurrency (Section~\ref{sec:limitations}).

\subsection{Robustness and calibration}
\label{sec:results-validation}

Appendix Tables~\ref{tab:validation} and~\ref{tab:app-heldout} collect further checks (donor-copy protocol). Risk K2 preserves 100.0\% of HotpotQA and 99.1\% of 2Wiki examples on which full prefill is exact-match correct, so preservation is not driven by reproducing wrong answers; paired bootstrap gains over random suffix K2 are +8.0 [5.6, 10.4] and +15.0 [11.8, 18.4] points; and held-out calibration on disjoint 30/70 splits (200 seeds) selects operating points without touching test rows.

\section{Discussion and Conclusion}

A cached RAG document is a context-conditioned causal state, and this paper shows what that means for reuse. At 7B scale, prefix-conditioned composition changes a large fraction of full-prefill answers on natural multi-hop RAG prompts (47.2\%/37.8\% preserved), and re-encoding the causal suffix from a selected start restores them (99.8\%/98.4\% under donor copy, 100.0\%/98.4\% under actual forwarding with risk K2). Document-only patching and fixed windows recover much less because a document's slice can be correct while every later slice keeps a history that omits it; the suffix is the smallest causally closed repair. Full-prefill answer equivalence is therefore the natural contract for non-prefix reuse, and SAFE-KV is a repair mechanism that meets it.

The same analysis maps the cost of meeting the contract. Reuse is exact for up to four documents, and at larger document counts the best repair unit depends on the budget. With training-free selectors, high-fidelity repair re-encodes most of the context, so in our batch-1 L40S study the operating rule serves every workload by full prefill. Because selector overhead is small and the cost is re-encoding, predicting a later start that still precedes the evidence is the lever that will turn faithful non-prefix reuse into a speed-up; SAFE-KV provides the contract, the mechanism, and the evaluation protocol against which such systems can be measured.

\label{page:mainend}
\section{Limitations}
\label{sec:limitations}

SAFE-KV establishes behavior preservation and measures its cost; it is not a serving system. Our latency study measures batch-1 time to first logit with one composed cache resident on a single L40S, including selector planning and excluding cache construction, lookup, transfer, batching, scheduling, and token emission. We did not integrate SAFE-KV with vLLM or another serving stack or measure throughput under concurrency, and no operating point we measured was both behavior-preserving and faster than full prefill.

The external evaluation uses a gold-informed construction: supporting documents come first and receive a construction-assigned retrieval score of 1.0, and the synthetic diagnostics likewise give evidence documents score 1.0 and the smallest document ids. The risk selector reads the retrieval score and both selectors break ties by ascending id, so selection on these constructions is easier than on real retrievers. The score-neutral shuffled runs (three seeds of 100 examples) are our label-free test of selection, and the query-selector rows of the synthetic diagnostics (the position sweep and the Mistral diagnostic) behave as oracle-start results (Section~\ref{sec:results-position}); on the external conversions the tie-break decides the start in at most 1.4\% of examples for the risk rows that carry the main claims. We did not evaluate real retriever traffic, so where evidence appears in production workloads remains open.

The donor-copy protocol inserts full-prefill slices, whereas the deployable protocol re-forwards the suffix from a composed prefix; the two agree closely on our inputs, and claims about deployable behavior rest on actual forwarding. The primary metric is a normalized short-answer exact match to full prefill after greedy decoding of at most 16 tokens; it does not measure raw-string, distributional, or hidden-state equality, it does not cover open-ended generation, and the full-prefill reference defines behavior rather than truth.

Our localized-repair comparison covers document-only patching, fixed downstream windows, linked intervals, and a query-attention token-local baseline that we implemented; we did not run Cache-Craft or CacheBlend. The experiments cover Qwen2.5-7B-Instruct and Mistral-7B-Instruct-v0.3 on document-level multi-hop QA; larger or proprietary models, long production traces, and multi-turn reuse are left to future work.

\section{Ethical Considerations}

SAFE-KV is an optimization-safety study over public QA benchmarks and synthetic diagnostics. We do not collect human-subject data, infer sensitive attributes, or release a user-facing model. The main ethical risk is deployment misuse: a non-prefix cache reuse system can silently change model behavior while appearing to return a normal answer. This is particularly important in RAG settings where users may rely on retrieved documents for current policy, legal, medical, financial, or operational decisions.

Non-prefix KV reuse should therefore be deployed only with explicit behavior-preservation checks, declared recomputation budgets, and fallback paths to full prefill for high-risk queries. SAFE-KV's full-prefill reference is not a truth oracle, and preserving full-prefill behavior does not guarantee factual correctness. Systems using SAFE-KV should still apply standard retrieval-quality checks, hallucination mitigation, privacy controls for cached documents, and access controls for any stored KV states derived from proprietary or personal data.

SAFE-KV's operating rule builds this safeguard in: a request is served from a repaired cache only when a validated operating point is both behavior-preserving and faster, and by full prefill otherwise, so speed is not bought with unvalidated behavior change. The experiments use substantial GPU computation (Appendix~\ref{sec:app-artifact} gives an estimate).

\bibliography{references}

\appendix
\setlength{\textfloatsep}{12pt plus 10pt minus 4pt}
\setlength{\dbltextfloatsep}{20pt plus 10pt minus 4pt}
\setlength{\floatsep}{12pt plus 6pt minus 2pt}
\setlength{\dblfloatsep}{12pt plus 6pt minus 2pt}
\setcounter{dbltopnumber}{2}         %
\setlength{\parskip}{0pt plus 2pt}    %
\section{Data Construction}
\label{sec:app-data}

\paragraph{External multi-hop data and the gold-informed construction.}
The main external evaluation uses HotpotQA-500 and 2Wiki-500. We remove yes/no questions, map each context paragraph to a SAFE-KV document, retain evidence labels, and keep distractor documents in the rendered prompt. The converter (\texttt{safekv/build\_external.py}, functions \texttt{\_select\_contexts} and \texttt{\_contexts\_to\_example}) emits the supporting paragraphs first, followed by distractors ranked by lexical overlap with the question, and assigns \texttt{retrieval\_score} $=1.0$ to supporting paragraphs and $\max(0.05,\,0.55-0.05\cdot i)$ to the $i$-th distractor. Document ids equal prompt positions. Consequently, in these files the evidence documents are always first, their retrieval scores are perfectly separated from the distractors' in 100\% of HotpotQA and 99.6\% of 2Wiki examples, and the risk selector's retrieval feature is gold-correlated. These files test behavior preservation under natural multi-document prompts in a controlled repair setting; selection without the construction cues is tested on the score-neutral variants below.

\paragraph{Score-neutral shuffled variants.}
\texttt{scripts/build\_rebuttal\_variants.py} rebuilds the same examples with every retrieval score set to $0.0$ and, for the shuffled variants, a deterministic per-example document shuffle with order seed 101, 202, or 303; document ids are reassigned to the new prompt positions, and evidence labels are carried along for analysis only. The first 100 examples of each variant are used in Table~\ref{tab:actual-forwarding}(b) and Table~\ref{tab:app-neutral-seed}. A clean CPU rebuild reproduces the saved bytes exactly.

\paragraph{Controlled diagnostics.}
The synthetic diagnostic data isolate failure modes: single-evidence lookup, multi-hop integration, conflict resolution, boundary-sensitive evidence, and distractor-heavy retrieval. Noisy variants preserve gold evidence while adding high-overlap distractors. In these files the generator assigns document ids before shuffling the prompt order, so ids do not equal prompt positions, evidence documents carry the smallest ids (0 or 1), and evidence documents receive retrieval score $1.0$ with distractors below $1.0$. Section~\ref{sec:selectors} and Appendix~\ref{sec:app-selector} describe how the selectors' ascending-id tie-break interacts with this.

\paragraph{Position sweep.}
The position-sensitivity benchmark fixes the answer-bearing document and moves it across first, early, middle, late, and last positions in a 19-document high-noise prompt while holding the distractor pool constant (50 examples per position). It tests whether local document repair is enough when downstream document states were computed without the evidence.

\paragraph{Staleness-by-document-count inputs.}
The nested single-evidence inputs \path{single_nested_n{1,2,4,8,16}_zero_layout101.jsonl} are built from the Noise16 pool with \texttt{--nested-single-evidence}, 50 paired examples shared across $N$, retrieval scores $0.0$, ids equal to prompt positions, and a label-independent evidence layout (seed 101). During evaluation each eligible document slice (position $\geq 2$) is replaced by its static-prefix encoding with probability $p$ under mask seeds 13, 29, and 47; the first document is always exact and the query is always recomputed.

\paragraph{Localized-baseline conversions.}
The operational token-local baseline uses separate HotpotQA distractor-setting conversions with about ten paragraphs per example in three orders: natural (source order), shuffled, and BM25-ranked (BM25 scores as retrieval scores; natural and shuffled scores $0.0$). Gold labels are not used for ranking. The mean one-based rank of the first supporting paragraph in the 500-example test splits is 5.39 (natural), 5.42 (shuffled), and 3.23 (BM25). These ranks describe our conversions rather than real RAG traffic. The first 100 test examples of each order are evaluated.

\paragraph{Full-prefill-correct subsets.}
Because the full-prefill behavioral reference can be gold-wrong, we also report subsets where full prefill exactly matches the gold answer (272 HotpotQA and 226 2Wiki examples).

\paragraph{Small-model development diagnostics.}
Earlier Qwen2.5-0.5B, Qwen2.5-1.5B, and SmolLM2-1.7B diagnostic tables document the development path and serve as sanity checks. The conclusions rest on the Qwen2.5-7B-Instruct runs, the Mistral-7B cross-model diagnostic, and the analyses in Sections~\ref{sec:results-main}--\ref{sec:results-validation}.

\section{Prompt Templates and Tokenization}
All answer-level experiments use the \texttt{chat\_rag} prompt format. The plain user content is:
\begin{verbatim}
[Document i | Title: <title>]
<document text>

Question: <question>
Answer with the shortest
exact answer phrase
from the documents.
Answer:
\end{verbatim}
This content is wrapped by the tokenizer's chat template with an added generation prompt. The implementation verifies that the chat template preserves the RAG content verbatim before shifting character ranges.

Token spans are computed by one-pass tokenization of the full rendered prompt with offset mappings. Document block ranges, document text ranges, evidence spans, boundary windows, and query suffix ranges are all derived from these offsets. This avoids retokenizing substrings independently and makes cache slicing consistent with the actual model input.

\paragraph{Why one-pass tokenization matters.}
KV cache slicing is sensitive to exact token boundaries. Retokenizing a document substring can produce a different tokenization at leading spaces, newlines, or chat-template boundaries, which would silently misalign cache slices. SAFE-KV instead tokenizes the full rendered prompt once, maps character offsets to token offsets, and stores all downstream ranges in that coordinate system. This is also why the prompt builder records both document-block ranges and document-text ranges: the block includes titles and delimiters, while text ranges support evidence-span diagnostics.

\paragraph{Answer instruction.}
The answer instruction asks for the shortest exact answer phrase from the documents. This keeps answer-level comparison focused: deterministic generations are short, normalization is meaningful, and changed answers usually correspond to evidence or source-selection changes rather than long-form stylistic variation. The instruction also reduces noise in full-prefill gold metrics, though it does not eliminate model mistakes on external data.

\paragraph{Stop behavior.}
Generation stops on newlines and on prompt-like markers such as \texttt{Human:}, \texttt{[Document}, \texttt{Question:}, and \texttt{Answer:}. These stops prevent pathological continuation into a new synthetic prompt and make answer equivalence easier to interpret. The same stops are used for full-prefill and composed-cache runs.

\section{Cache Mechanics Sanity Checks}
The implementation includes generation sanity checks comparing manual cached decoding with Hugging Face generation. The matching mode uses manual position IDs. Sequential prefill sanity verifies that splitting the prompt into pieces and feeding them in order recovers full-prefill behavior. Full-document patch sanity replaces all document cache slices with full-prefill slices and recovers the full-prefill first-token behavior, validating the patching mechanism.

These checks are important because the central paper claim is semantic rather than mechanical. If manual decoding, position IDs, or cache slicing were wrong, drift could be explained by an implementation artifact. The sanity suite instead narrows the interpretation: full prefill, sequential prefill, and full-document patching agree when they should, so the remaining discrepancies arise from the conditioning assumptions made by non-prefix composition.

\paragraph{Position IDs and cache length.}
All answer-level experiments use the validated manual-position-id path. This is important for non-prefix composition because the cache can be assembled from pieces whose original local positions differ from their positions in the final prompt. SAFE-KV verifies expected cache lengths before decoding and treats length mismatches as errors rather than silently continuing. This guardrail prevents accidental comparisons between prompts with different effective contexts.

\paragraph{Full-document patch sanity.}
The full-document patch sanity condition is the most direct check of repair mechanics. Starting from a composed cache, the implementation replaces every document slice with the corresponding full-prefill slice. If patching is correct, the resulting first-token behavior should match full prefill. The observed near-zero KL for this sanity method establishes that document slices can be patched without corrupting the rest of the cache, so weaker span/window results reflect repair granularity rather than a broken patch operator.

\paragraph{Sequential prefill sanity.}
Sequential prefill is a second control. The prompt is split and fed through the model in order while preserving the past-key-value state. Agreement with one-shot full prefill confirms that the generation wrapper, attention mask, and position-id handling can reproduce the model's normal causal computation when the causal history is correct.

\section{Algorithms}
\label{sec:app-algorithms}

\paragraph{Algorithm 0: full-prefill behavioral reference.}
The reference is the model's ordinary causal computation on the exact rendered RAG prompt.
\begin{enumerate}
    \item Render the full chat/RAG prompt with prefix $p$, ordered documents $d_1,\ldots,d_m$, and query suffix $q$.
    \item Tokenize the complete prompt once and record prefix, document, and query token ranges in global coordinates.
    \item Run one full-prefill forward pass with cache enabled.
    \item Store the first-answer-token distribution, the greedy generated answer, and (for the donor-copy diagnostic) per-document cache slices.
    \item Treat this behavior as the reference for answer preservation, changed-wrong counts, and first-token KL.
\end{enumerate}

\paragraph{Algorithm 1: prefix-conditioned composition.}
\begin{enumerate}
    \item Build the static prefix cache for $p$.
    \item For each document $d_i$, prefill $(p,d_i)$ with $d_i$ at its target global position and extract the cache slice aligned to $d_i$.
    \item Concatenate prefix and document slices into a composed context cache $C_{\mathrm{comp}}$.
    \item Evaluate the query suffix against the composed cache with manual position IDs.
    \item Decode greedily and compare with full prefill.
\end{enumerate}

\paragraph{Algorithm 2: suffix repair under the two protocols.}
Given selected ids $S$ and $s=\min\{i:d_i\in S\}$:
\begin{enumerate}
    \item Build the prefix-conditioned composed cache and score documents with the query, risk, or hybrid selector (Section~\ref{sec:selectors}).
    \item \textbf{Actual forwarding} (deployable; \path{run_answer_eval.py::recompute_answer_suffix}, methods \texttt{actual\_recompute::*}): slice the composed cache to positions $[0, \mathrm{start}(d_s))$, run the contiguous token range $[\mathrm{start}(d_s), \mathrm{start}(q))$ through the model with that prefix cache and its global position ids, and replace the composed slices for $d_s,\ldots,d_m$ with the resulting key/value states. The runner rejects ranges with gaps, ranges that do not end at the query boundary, and ranges that start before the first document.
    \item \textbf{Donor copy} (offline diagnostic): replace the composed slices for $d_s,\ldots,d_m$ with the corresponding slices of the separately computed full-prefill cache.
    \item Encode the query against the repaired cache and decode greedily.
\end{enumerate}
The two protocols evaluate different states: donor copy inserts slices computed under the exact full history, actual forwarding computes them under the retained composed prefix. Table~\ref{tab:app-recon} shows that they agree closely on the tested inputs. Every latency measurement uses actual forwarding.

\paragraph{Algorithm 3: local and window ablations.}
\begin{enumerate}
    \item Select a risk or oracle document $d_s$.
    \item For document-only repair, re-encode or replace only $d_s$.
    \item For window repair with size $w$, re-encode or replace $d_s,\ldots,d_{\min(s+w-1,m)}$.
    \item Leave later documents outside the window in their composed-cache state.
    \item Decode and compare with suffix repair at similar budgets.
\end{enumerate}

\paragraph{Algorithm 4: held-out operating-point calibration.}
\begin{enumerate}
    \item Split examples into 30\% calibration and 70\% test using deterministic seeds (150/350 of 500; the two sets are disjoint by construction).
    \item On calibration examples, choose the lowest-budget operating point whose answer-preservation estimate meets the target.
    \item Apply that operating point unchanged to the held-out test examples.
    \item Repeat over 200 split seeds and report test answer preservation, changed-wrong rate, and recomputed ratio.
\end{enumerate}

\paragraph{Algorithm 5: query-attention token-local baseline.}
\begin{enumerate}
    \item Run the query against the stale composed cache with an FP32 eager attention pass and average the query-to-context attention over layers, heads, and query positions.
    \item Rank all document-block tokens by that score (ties by position) and select the top $\lfloor b\cdot n\rfloor$ tokens for a document-token budget $b$; merge them into contiguous ranges.
    \item Re-encode each selected range in ascending order by actual forwarding under the cache state that precedes it (including earlier repaired ranges); unselected gaps and all later states remain composed.
    \item Encode the query against the repaired cache and decode greedily. Selection reads no gold label, full-prefill state, or mismatch mask (our implementation of a selective-recomputation baseline).
\end{enumerate}

\section{Composition Ablation Diagnostics}
Table~\ref{tab:app-composition} reports the early composition variants that identified prefix-conditioned composition as the strongest simple non-prefix baseline (first-token diagnostics on a small model).
\begin{table*}[t]
\centering
\footnotesize
\begin{adjustbox}{max width=\textwidth}
\begin{tabular}{ccccc}
\toprule
method & avg kl & top1 & top10 & patch ratio \\
\midrule
Doc-only & 0.85 & 0.61 & 0.56 & 0.00 \\
Doc-only local & 1.48 & 0.45 & 0.57 & 0.00 \\
Prefix-cond. & 0.21 & 0.90 & 0.70 & 0.00 \\
Prefix-cond. local & 0.29 & 0.80 & 0.69 & 0.00 \\
Prefix-plus & 1.21 & 0.47 & 0.57 & 0.00 \\
Prefix-plus local & 1.88 & 0.36 & 0.55 & 0.00 \\
\bottomrule
\end{tabular}
\end{adjustbox}
\caption{Full early composition ablation (Qwen2.5-0.5B-Instruct, clean diagnostic set, 100 examples): first-token KL, top-1 and top-10 agreement with full prefill for doc-only, prefix-conditioned, and prefix-plus composition with global or local position ids.}
\label{tab:app-composition}
\end{table*}

\section{Early Repair-Granularity Diagnostics}
Table~\ref{tab:app-repair} reports the early repair-granularity diagnostic from method development: first-token metrics on Qwen2.5-0.5B for sub-document units (evidence spans, span windows, and span-to-document-end regions). These units did not separate from matched random patching, which led the method to document-level and then document-suffix units evaluated at the answer level on 7B models (Sections~\ref{sec:results-main}--\ref{sec:results-position}); the table does not contain the document-suffix repair of Section~\ref{sec:problem}.
\begin{table*}[t]
\centering
\footnotesize
\begin{adjustbox}{max width=\textwidth}
\begin{tabular}{ccccc}
\toprule
method & avg kl & top1 & top10 & patch ratio \\
\midrule
Oracle doc & 0.09 & 0.97 & 0.77 & 0.46 \\
Sent. to doc end & 0.23 & 0.91 & 0.71 & 0.19 \\
Span & 0.24 & 0.91 & 0.71 & 0.11 \\
Span+16 & 0.20 & 0.92 & 0.72 & 0.24 \\
Span+32 & 0.20 & 0.92 & 0.72 & 0.30 \\
Span+4 & 0.23 & 0.91 & 0.71 & 0.14 \\
Span+8 & 0.22 & 0.91 & 0.71 & 0.18 \\
Evidence suffix & 0.23 & 0.91 & 0.71 & 0.19 \\
S+C span & 0.19 & 0.95 & 0.72 & 0.13 \\
S+C suffix & 0.18 & 0.96 & 0.72 & 0.22 \\
Rand sent-doc & 0.18 & 0.92 & 0.71 & 0.19 \\
Rand span & 0.20 & 0.92 & 0.70 & 0.11 \\
Rand span+16 & 0.18 & 0.94 & 0.71 & 0.24 \\
Rand span+32 & 0.16 & 0.94 & 0.73 & 0.30 \\
Rand span+4 & 0.18 & 0.93 & 0.71 & 0.14 \\
Rand span+8 & 0.19 & 0.93 & 0.70 & 0.18 \\
Rand suffix & 0.18 & 0.94 & 0.71 & 0.19 \\
Rand S+C span & 0.17 & 0.93 & 0.71 & 0.13 \\
Matched random & 0.15 & 0.95 & 0.72 & 0.22 \\
\bottomrule
\end{tabular}
\end{adjustbox}
\caption{Full early repair-granularity diagnostic (Qwen2.5-0.5B-Instruct, clean diagnostic set, 100 examples, donor-copy protocol): first-token KL, top-1 and top-10 agreement, and patch ratio for sub-document units (evidence spans, span windows of 4--32 tokens, sentence- or span-to-document-end regions, labelled ``suffix'' here but internal to one document) and their matched random baselines. These first-token metrics do not separate the sub-document units from random patching; the document-suffix repair evaluated at 7B (Section~\ref{sec:problem}) is a different unit and is not represented in this table.}
\label{tab:app-repair}
\end{table*}

\section{Small-Model Development Tables}
Table~\ref{tab:app-synthetic} reports earlier clean/noisy synthetic answer-level runs on smaller open models. These development runs show the same qualitative pattern of prefix-composition drift and repair sensitivity that the 7B-scale experiments establish.
\begin{table*}[t]
\centering
\footnotesize
\begin{adjustbox}{max width=\textwidth}
\begin{tabular}{cccccc}
\toprule
model & split & method & answer equivalence & changed wrong & patch ratio \\
\midrule
qwen05 & clean & Entity K2 & 0.97 & 1 & 0.54 \\
qwen05 & clean & Full & 1.00 & 0 & 0.00 \\
qwen05 & clean & Oracle doc & 0.97 & 1 & 0.46 \\
qwen05 & clean & Prefix-cond. & 0.85 & 14 & 0.00 \\
qwen05 & clean & Query K2 & 0.97 & 1 & 0.54 \\
qwen05 & clean & Random K2 & 0.96 & 3 & 0.53 \\
qwen05 & clean & Risk K2 & 0.97 & 1 & 0.55 \\
qwen05 & clean & Gate .45 & 0.98 & 1 & 0.36 \\
qwen05 & clean & Gate .60 & 0.92 & 7 & 0.11 \\
qwen05 & noisy & Entity K2 & 0.90 & 7 & 0.55 \\
qwen05 & noisy & Full & 1.00 & 0 & 0.00 \\
qwen05 & noisy & Oracle doc & 0.90 & 6 & 0.42 \\
qwen05 & noisy & Prefix-cond. & 0.69 & 23 & 0.00 \\
qwen05 & noisy & Query K2 & 0.89 & 7 & 0.55 \\
qwen05 & noisy & Random K2 & 0.83 & 11 & 0.54 \\
qwen05 & noisy & Risk K2 & 0.86 & 11 & 0.55 \\
qwen05 & noisy & Gate .45 & 0.86 & 10 & 0.54 \\
qwen05 & noisy & Gate .60 & 0.81 & 14 & 0.28 \\
qwen15 & clean & Entity K2 & 0.96 & 3 & 0.54 \\
qwen15 & clean & Full & 1.00 & 0 & 0.00 \\
qwen15 & clean & Oracle doc & 0.96 & 3 & 0.46 \\
qwen15 & clean & Prefix-cond. & 0.88 & 11 & 0.00 \\
qwen15 & clean & Query K2 & 0.96 & 3 & 0.54 \\
qwen15 & clean & Random K2 & 0.93 & 6 & 0.53 \\
qwen15 & clean & Gate .45 & 0.96 & 3 & 0.36 \\
qwen15 & clean & Gate .60 & 0.92 & 6 & 0.11 \\
qwen15 & noisy & Entity K2 & 0.92 & 6 & 0.55 \\
qwen15 & noisy & Full & 1.00 & 0 & 0.00 \\
qwen15 & noisy & Oracle doc & 0.89 & 10 & 0.42 \\
qwen15 & noisy & Prefix-cond. & 0.74 & 20 & 0.00 \\
qwen15 & noisy & Query K2 & 0.92 & 6 & 0.55 \\
qwen15 & noisy & Random K2 & 0.86 & 8 & 0.54 \\
qwen15 & noisy & Gate .45 & 0.92 & 6 & 0.54 \\
qwen15 & noisy & Gate .60 & 0.84 & 13 & 0.28 \\
\bottomrule
\end{tabular}
\end{adjustbox}
\caption{Full synthetic answer-level table.}
\label{tab:app-synthetic}
\end{table*}

\section{External, Cross-Model, and Protocol Details}
\label{sec:app-external}

Table~\ref{tab:cross-external} summarizes the external 7B evidence and the Mistral-7B noisy diagnostic check under the donor-copy protocol; the full external table is repeated in Table~\ref{tab:app-external}.

\begin{table*}[t]
\centering
\footnotesize
\begin{adjustbox}{max width=\textwidth}
\begin{tabular}{llrrrr}
\toprule
Setting & Method & Prefix eq. & Best SAFE-KV eq. & Random eq. & Patch \\
\midrule
HotpotQA-500, Qwen2.5-7B & Risk suffix K2 & 0.472 & 0.998 & 0.918 & 0.948 \\
2Wiki-500, Qwen2.5-7B & Risk suffix K2 & 0.378 & 0.984 & 0.834 & 0.929 \\
2Wiki exact-correct, Qwen2.5-7B & Risk suffix K2 & 0.518 & 0.991 & 0.885 & 0.921 \\
Hotpot exact-correct, Qwen2.5-7B & Risk suffix K2 & 0.566 & 1.000 & 0.949 & 0.949 \\
Noisy-250, Mistral-7B & Query/Risk suffix K2 & 0.312 & 0.996 & 1.000 & 0.872--0.879 \\
\bottomrule
\end{tabular}
\end{adjustbox}
\caption{External and cross-model summary. Prefix eq. is answer equivalence of prefix-conditioned composition; Random eq. is matched random suffix K2 when available. Patch is the reported SAFE-KV method's patched-token ratio, computed on the same subset for exact-correct rows.}
\label{tab:cross-external}
\end{table*}

\begin{table*}[t]
\centering
\footnotesize
\begin{adjustbox}{max width=\textwidth}
\begin{tabular}{llrrrrr}
\toprule
Data & Method & Ans. eq. & Gold EM & Gold F1 & Changed-wrong & Patch \\
\midrule
HotpotQA-500 & Full prefill & 1.000 & 0.544 & 0.708 & 0 & 0.000 \\
HotpotQA-500 & Prefix-comp. & 0.472 & 0.354 & 0.489 & 241 & 0.000 \\
HotpotQA-500 & Query suffix K1 & 0.930 & 0.522 & 0.696 & 32 & 0.842 \\
HotpotQA-500 & Query suffix K2 & 0.958 & 0.534 & 0.704 & 18 & 0.883 \\
HotpotQA-500 & Random suffix K2 & 0.918 & 0.528 & 0.696 & 35 & 0.814 \\
HotpotQA-500 & Risk suffix K1 & 0.996 & 0.542 & 0.708 & 2 & 0.892 \\
HotpotQA-500 & Hybrid suffix K1 & 0.996 & 0.542 & 0.708 & 2 & 0.904 \\
HotpotQA-500 & Risk suffix K2 & 0.998 & 0.544 & 0.709 & 1 & 0.948 \\
\midrule
2Wiki-500 & Full prefill & 1.000 & 0.452 & 0.546 & 0 & 0.000 \\
2Wiki-500 & Prefix-comp. & 0.378 & 0.312 & 0.386 & 272 & 0.000 \\
2Wiki-500 & Query suffix K2 & 0.972 & 0.444 & 0.538 & 12 & 0.885 \\
2Wiki-500 & Random suffix K2 & 0.834 & 0.434 & 0.517 & 66 & 0.711 \\
2Wiki-500 & Risk suffix K1 & 0.970 & 0.442 & 0.536 & 12 & 0.813 \\
2Wiki-500 & Margin-adaptive & 0.972 & 0.440 & 0.534 & 12 & 0.838 \\
2Wiki-500 & Hybrid suffix K1 & 0.974 & 0.442 & 0.536 & 11 & 0.842 \\
2Wiki-500 & Risk suffix K2 & 0.984 & 0.452 & 0.546 & 6 & 0.929 \\
\bottomrule
\end{tabular}
\end{adjustbox}
\caption{Full Qwen2.5-7B external answer-level results.}
\label{tab:app-external}
\end{table*}

\begin{figure*}[t]
\centering
\includegraphics[width=0.95\linewidth]{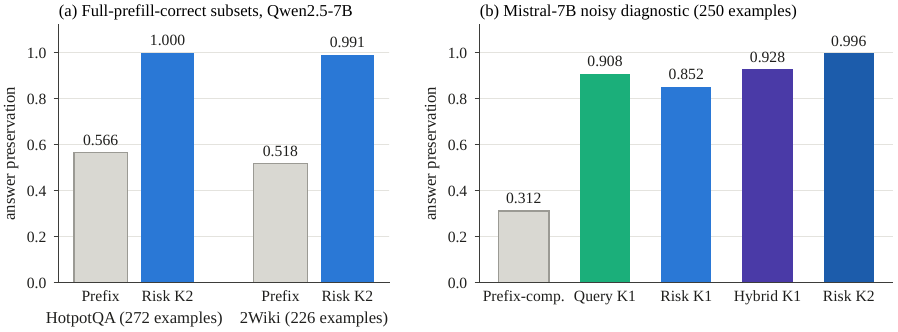}
\caption{Robustness checks (donor-copy protocol; plotted values in \texttt{figure5\allowbreak\_external\allowbreak\_sanity.data.csv}). Left: on examples where full prefill is exact-match correct (HotpotQA 272, 2Wiki 226), risk suffix K2 preserves nearly all correct full-prefill behavior. Right: the Mistral-7B noisy diagnostic (250 examples) shows the same unsafe-composition failure and suffix-repair recovery pattern; its query and hybrid rows are largely decided by the id tie-break (Section~\ref{sec:results-selector}).}
\label{fig:app-robustness}
\end{figure*}

\paragraph{Interpreting external gold scores.}
External gold metrics are secondary because SAFE-KV evaluates whether cache reuse changes the model relative to full prefill. A method can preserve a full-prefill answer that is gold-wrong, and a method can improve gold EM by changing the model's behavior. The primary metric is therefore answer preservation, while gold EM/F1 diagnose whether preservation tracks benchmark quality.

\paragraph{Full-prefill-correct subset analysis.}
On HotpotQA exact-correct examples (272; Figure~\ref{fig:app-robustness}, left), risk K1 and risk K2 reach 99.6\% and 100.0\% answer preservation; on 2Wiki exact-correct examples (226), risk K1, hybrid K1, and risk K2 reach 96.5\%, 96.9\%, and 99.1\%. Behavior preservation therefore also preserves the answers that full prefill gets right.

\paragraph{Why random suffix is strong but not sufficient.}
Random suffix repair re-encodes a coherent downstream region, so it is a strong baseline. On the main external runs the selected suffixes are materially stronger: risk K2 improves over random K2 by 8.0 points on HotpotQA and 15.0 on 2Wiki in paired bootstrap analysis. On the Mistral noisy diagnostic random suffix K2 also reaches 100.0\%: there, any sufficiently early start suffices, and the result reflects the repair mechanism rather than the selector.

\paragraph{Actual forwarding: per-seed values and same-run reconciliation.}
Table~\ref{tab:app-neutral-seed} lists the per-seed values behind Table~\ref{tab:actual-forwarding}(b). Table~\ref{tab:app-recon} reports a same-run comparison of donor copy and actual forwarding on the original inputs: on HotpotQA every generated answer is identical under the two protocols; on 2Wiki they differ on 17 (K1) and 7 (K2) of 500 outputs, and actual forwarding is the lower of the two.

\begin{table*}[t]
\centering
\footnotesize
\begin{adjustbox}{max width=\textwidth}
\begin{tabular}{llrrr}
\toprule
Data & Method & seed 101 & seed 202 & seed 303 \\
\midrule
HotpotQA & Risk suffix K1 & 0.81 / 0.78 / 0.579 & 0.85 / 0.85 / 0.564 & 0.83 / 0.83 / 0.549 \\
HotpotQA & Hybrid suffix K1 & 0.92 / 0.90 / 0.739 & 0.91 / 0.91 / 0.721 & 0.90 / 0.90 / 0.712 \\
HotpotQA & Query suffix K2 & 0.97 / 0.96 / 0.880 & 0.97 / 0.97 / 0.869 & 0.94 / 0.94 / 0.847 \\
HotpotQA & Risk suffix K2 & 0.90 / 0.89 / 0.780 & 0.92 / 0.92 / 0.751 & 0.93 / 0.93 / 0.755 \\
\midrule
2Wiki & Risk suffix K1 & 0.72 / 0.71 / 0.622 & 0.79 / 0.77 / 0.612 & 0.81 / 0.79 / 0.632 \\
2Wiki & Hybrid suffix K1 & 0.81 / 0.80 / 0.710 & 0.88 / 0.87 / 0.733 & 0.84 / 0.82 / 0.698 \\
2Wiki & Query suffix K2 & 0.93 / 0.92 / 0.854 & 0.96 / 0.95 / 0.870 & 0.97 / 0.97 / 0.867 \\
2Wiki & Risk suffix K2 & 0.89 / 0.88 / 0.801 & 0.91 / 0.89 / 0.798 & 0.93 / 0.92 / 0.803 \\
\bottomrule
\end{tabular}
\end{adjustbox}
\caption{Per-seed values behind Table~\ref{tab:actual-forwarding}(b): normalized answer preservation / generated-token exact / mean recomputed context fraction for each score-neutral shuffled order seed (100 examples each, actual forwarding, one NVIDIA L40S).}
\label{tab:app-neutral-seed}
\end{table*}

\begin{table*}[t]
\centering
\footnotesize
\begin{adjustbox}{max width=\textwidth}
\begin{tabular}{llrrrrr}
\toprule
Data & Method & Donor ans.\ eq. & Actual ans.\ eq. & Same generated answer & Donor token exact & Actual token exact \\
\midrule
HotpotQA-500 & Risk suffix K1 & 0.998 & 0.998 & 1.000 & 0.998 & 0.998 \\
HotpotQA-500 & Risk suffix K2 & 1.000 & 1.000 & 1.000 & 1.000 & 1.000 \\
2Wiki-500 & Risk suffix K1 & 0.978 & 0.966 & 0.966 & 0.972 & 0.958 \\
2Wiki-500 & Risk suffix K2 & 0.992 & 0.984 & 0.986 & 0.990 & 0.980 \\
\bottomrule
\end{tabular}
\end{adjustbox}
\caption{Same-run donor-copy versus actual-forwarding reconciliation on the original gold-evidence-first inputs (500 examples each; identical selected documents and patch ratios in every row; validator passed). This rerun used PyTorch~2.8.0+cu128 on one NVIDIA L40S, whereas Table~\ref{tab:external-main} was produced under PyTorch~2.4.1+cu124; the donor-copy 2Wiki values here (0.978/0.992) therefore differ slightly from the 0.970/0.984 of Table~\ref{tab:external-main} although inputs, selections and patch counts are byte-identical. The difference traces to FP16/SDPA near-tie sensitivity across runtime versions; code, data, and selections are unchanged.}
\label{tab:app-recon}
\end{table*}

\section{Robustness and Calibration Checks}

Table~\ref{tab:validation} collects compact validation checks for common pitfalls in cache-reuse evaluation: gains over random suffix controls, held-out method calibration rather than test-set threshold tuning, and measured first-token latency rather than patch-ratio-only accounting.

\begin{table*}[t]
\centering
\footnotesize
\begin{adjustbox}{max width=\textwidth}
\begin{tabular}{llllll}
\toprule
Check & Setting & Value & CI / Split interval & Error / sample detail & Patch / latency detail \\
\midrule
Bootstrap & Hotpot risk K2 $-$ random K2 & +0.080 & [0.056, 0.104] & $\Delta$ changed-wrong -0.068 & $\Delta$ patch +0.134 \\
Bootstrap & 2Wiki risk K2 $-$ random K2 & +0.150 & [0.118, 0.184] & $\Delta$ changed-wrong -0.120 & $\Delta$ patch +0.218 \\
Bootstrap & 2Wiki hybrid K1 $-$ risk K1 & +0.004 & [0.000, 0.010] & $\Delta$ changed-wrong -0.002 & $\Delta$ patch +0.029 \\
Calibration & 2Wiki target 0.85 & 0.970 & [0.960, 0.980] & changed-wrong 0.024 & patch 0.824 \\
Calibration & 2Wiki target 0.94 & 0.979 & [0.960, 0.989] & changed-wrong 0.016 & patch 0.908 \\
Latency & 2Wiki-50 full / risk K1 & 0.071s / 0.083s & p95 0.115s / 0.127s & n=50 & patch 0.814 \\
Latency & noise64-20 full / risk K1 & 0.215s / 0.140s & p95 0.224s / 0.236s & n=20 & patch 0.481 \\
\bottomrule
\end{tabular}
\end{adjustbox}
\caption{Robustness and calibration checks (donor-copy protocol, gold-evidence-first construction). Bootstrap deltas are paired over examples; calibration selects methods on 30\% of examples and reports held-out 70\% performance over 200 splits. The two latency rows are early prototype timings (donor-copy suffix, no selector planning; Table~\ref{tab:app-cost}); the paper's latency conclusions use the actual-forwarding measurements of Table~\ref{tab:latency-main}.}
\label{tab:validation}
\end{table*}

Together with the held-out calibration and latency appendices, these checks make the main claims auditable: SAFE-KV improves over a strong matched random suffix baseline on the main external runs, retains strong held-out behavior under calibration splits, and is timed separately from its token budget.

\section{Held-Out Calibration Details}

Held-out calibration chooses among SAFE-KV operating points on calibration examples and evaluates the chosen policy on held-out examples. The purpose is to avoid selecting a threshold or policy after inspecting the test rows. Table~\ref{tab:app-heldout} reports representative target levels over 200 deterministic split seeds; each split assigns 150 examples to calibration and the remaining 350 to test, so calibration and test rows never overlap. All values use the donor-copy protocol on the gold-evidence-first construction.

\begin{table*}[t]
\centering
\footnotesize
\begin{adjustbox}{max width=\textwidth}
\begin{tabular}{llrrrr}
\toprule
Data & Target & Test ans. eq. & Split interval & Changed-wrong & Patch \\
\midrule
HotpotQA-500 & 0.82 & 0.996 & [0.994, 1.000] & 0.004 & 0.905 \\
HotpotQA-500 & 0.89 & 0.980 & [0.951, 1.000] & 0.018 & 0.891 \\
HotpotQA-500 & 0.95 & 0.996 & [0.994, 1.000] & 0.004 & 0.901 \\
\midrule
2Wiki-500 & 0.82 & 0.973 & [0.963, 0.989] & 0.022 & 0.842 \\
2Wiki-500 & 0.85 & 0.970 & [0.960, 0.980] & 0.024 & 0.824 \\
2Wiki-500 & 0.89 & 0.970 & [0.960, 0.977] & 0.025 & 0.842 \\
2Wiki-500 & 0.94 & 0.979 & [0.960, 0.989] & 0.016 & 0.908 \\
\bottomrule
\end{tabular}
\end{adjustbox}
\caption{Held-out SAFE-KV calibration over 30\%/70\% calibration/test splits. Targets choose among SAFE-KV operating points on calibration examples; all reported metrics are held-out test metrics.}
\label{tab:app-heldout}
\end{table*}

Calibrated policies trade a little fidelity for a declared budget, so they need not match the best individual method in Table~\ref{tab:external-main}; they show that SAFE-KV can fix a target reliability/budget region before evaluation and still preserve most full-prefill behavior on held-out examples.

\section{Latency and Cost Accounting}
\label{sec:app-cost}

Tables~\ref{tab:app-lat-full} and~\ref{tab:app-plan} and the Noise64 gate in Table~\ref{tab:app-noise64-gate} are the measurements behind Section~\ref{sec:results-latency}: one NVIDIA L40S, actual suffix forwarding with selector planning inside the timed region, five repeats per example, and an independent validator. Table~\ref{tab:app-cost} lists earlier prototype timings with a narrower scope: donor-copied suffix slices, no selector scoring or suffix-start planning in the timed path, and one timing per example on an L40S-class GPU. The two differ in measurement scope rather than hardware; the main text uses the former.

\begin{table*}[t]
\centering
\footnotesize
\begin{adjustbox}{max width=\textwidth}
\begin{tabular}{llrrrrr}
\toprule
Data & Method & Avg TTFT & Median & p95 & Peak MB & Patch \\
\midrule
2Wiki-50 & Full prefill & 0.071 & 0.060 & 0.115 & 15204 & 0.000 \\
2Wiki-50 & Prefix query forward & 0.026 & 0.025 & 0.029 & 15038 & 0.000 \\
2Wiki-50 & Query suffix K1 & 0.079 & 0.071 & 0.121 & 15180 & 0.732 \\
2Wiki-50 & Risk suffix K1 & 0.083 & 0.078 & 0.127 & 15198 & 0.814 \\
2Wiki-50 & Hybrid suffix K1 & 0.084 & 0.080 & 0.126 & 15202 & 0.834 \\
2Wiki-50 & Risk suffix K2 & 0.089 & 0.087 & 0.131 & 15223 & 0.940 \\
\midrule
Noise64-20 & Full prefill & 0.215 & 0.192 & 0.224 & 16545 & 0.000 \\
Noise64-20 & Prefix query forward & 0.027 & 0.027 & 0.029 & 16021 & 0.000 \\
Noise64-20 & Query suffix K1 & 0.162 & 0.167 & 0.236 & 16428 & 0.597 \\
Noise64-20 & Risk suffix K1 & 0.140 & 0.131 & 0.236 & 16345 & 0.481 \\
Noise64-20 & Hybrid suffix K1 & 0.163 & 0.167 & 0.235 & 16430 & 0.600 \\
Noise64-20 & Risk suffix K2 & 0.176 & 0.186 & 0.237 & 16474 & 0.666 \\
\bottomrule
\end{tabular}
\end{adjustbox}
\caption{Early prototype first-token latency (seconds) and patch ratios. Protocol: donor-copy suffix slices plus query forward; selector scoring and suffix-start planning outside the timed path; one timing per example; L40S-class GPU (exact model not recorded). Peak MB is average peak CUDA memory over examples. Prefix query forward excludes the cost of building reusable context caches and is shown only to separate query evaluation from repair work. The paper's latency conclusions use the actual-forwarding measurements of Tables~\ref{tab:latency-main} and~\ref{tab:app-lat-full}.}
\label{tab:app-cost}
\end{table*}

\begin{table*}[t]
\centering
\footnotesize
\begin{adjustbox}{max width=\textwidth}
\begin{tabular}{lrrrrrr}
\toprule
Method & Median (ms) & p95 (ms) & Repair ratio & Incr.\ peak MB & Paired speedup [95\% CI] & Faster \\
\midrule
\multicolumn{7}{l}{\emph{HotpotQA, original order} (mean prompt 759.8 tokens)} \\
Full prefill & 65.5 & 88.7 & 1.000 & 146 & --- & 0\% \\
Prefix query only (no repair; diagnostic) & 34.3 & 35.8 & 0.000 & 55 & 1.903 [1.824, 1.970] & 100\% \\
Query suffix K1 & 91.3 & 121.2 & 0.852 & 123 & 0.682 [0.671, 0.717] & 0\% \\
Risk suffix K1 & 93.6 & 129.6 & 0.917 & 130 & 0.655 [0.624, 0.675] & 0\% \\
Hybrid suffix K1 & 94.4 & 129.8 & 0.924 & 131 & 0.655 [0.622, 0.677] & 0\% \\
Risk suffix K2 & 104.2 & 132.7 & 0.952 & 134 & 0.629 [0.621, 0.670] & 0\% \\
\midrule
\multicolumn{7}{l}{\emph{2Wiki, original order} (mean prompt 775.1 tokens)} \\
Full prefill & 65.9 & 118.9 & 1.000 & 149 & --- & 0\% \\
Prefix query only (no repair; diagnostic) & 34.6 & 36.3 & 0.000 & 56 & 1.928 [1.798, 2.040] & 100\% \\
Query suffix K1 & 86.4 & 144.7 & 0.732 & 121 & 0.713 [0.669, 0.738] & 0\% \\
Risk suffix K1 & 88.5 & 147.1 & 0.814 & 127 & 0.667 [0.647, 0.704] & 0\% \\
Hybrid suffix K1 & 91.6 & 145.5 & 0.834 & 129 & 0.664 [0.633, 0.699] & 0\% \\
Risk suffix K2 & 105.4 & 163.6 & 0.940 & 137 & 0.629 [0.618, 0.668] & 0\% \\
\midrule
\multicolumn{7}{l}{\emph{2Wiki, score-neutral shuffled} (mean prompt 775.1 tokens)} \\
Full prefill & 65.6 & 119.0 & 1.000 & 149 & --- & 0\% \\
Prefix query only (no repair; diagnostic) & 34.4 & 36.3 & 0.000 & 56 & 1.912 [1.814, 2.049] & 100\% \\
Query suffix K1 & 85.8 & 145.2 & 0.690 & 118 & 0.691 [0.643, 0.741] & 4\% \\
Risk suffix K1 & 78.1 & 147.1 & 0.586 & 113 & 0.722 [0.678, 0.811] & 10\% \\
Hybrid suffix K1 & 86.7 & 146.5 & 0.697 & 118 & 0.686 [0.636, 0.739] & 4\% \\
Risk suffix K2 & 88.7 & 146.4 & 0.786 & 125 & 0.668 [0.625, 0.710] & 2\% \\
\midrule
\multicolumn{7}{l}{\emph{$N{=}16$ nested, $p{=}1$} (mean prompt 661.0 tokens)} \\
Full prefill & 62.1 & 63.0 & 1.000 & 127 & --- & 0\% \\
Prefix query only (no repair; diagnostic) & 32.4 & 33.0 & 0.000 & 46 & 1.920 [1.912, 1.926] & 100\% \\
Query suffix K1 & 89.2 & 90.6 & 0.950 & 117 & 0.695 [0.694, 0.697] & 0\% \\
Risk suffix K1 & 87.8 & 90.9 & 0.723 & 103 & 0.702 [0.692, 0.726] & 0\% \\
Hybrid suffix K1 & 89.9 & 91.5 & 0.950 & 117 & 0.689 [0.688, 0.691] & 0\% \\
Risk suffix K2 & 89.5 & 90.9 & 0.832 & 109 & 0.692 [0.690, 0.695] & 0\% \\
\midrule
\multicolumn{7}{l}{\emph{Noise64 (67 documents)} (mean prompt 2540.5 tokens)} \\
Full prefill & 199.4 & 219.4 & 1.000 & 487 & --- & 0\% \\
Prefix query only (no repair; diagnostic) & 36.2 & 38.9 & 0.000 & 176 & 5.516 [5.470, 5.584] & 100\% \\
Query suffix K1 & 184.7 & 258.5 & 0.554 & 371 & 1.120 [0.995, 1.467] & 64\% \\
Risk suffix K1 & 174.6 & 260.6 & 0.519 & 360 & 1.149 [1.027, 1.426] & 72\% \\
Hybrid suffix K1 & 192.7 & 262.1 & 0.593 & 377 & 1.027 [0.935, 1.258] & 58\% \\
Risk suffix K2 & 203.8 & 267.4 & 0.659 & 387 & 0.977 [0.896, 1.087] & 48\% \\
\bottomrule
\end{tabular}
\end{adjustbox}
\caption{All timed methods for the five L40S workloads of Table~\ref{tab:latency-main} (actual forwarding; selector scoring and suffix-start planning inside the timed region for the four SAFE-KV rows). ``Prefix query only'' evaluates the query against the unrepaired composed cache and is shown only to separate query evaluation from repair work; it is not an operating point. Incr.\ peak MB is the incremental peak CUDA memory of the request above the resident model and cache pool. Each run passed the independent validator (1{,}500 latency rows, 200 selector rows, 50 cache-build rows, zero failures). One-shot composed-cache construction, outside the request path, had median 201.5 / 204.4 / 203.1 / 532.3 / 2174.0~ms for the five workloads.}
\label{tab:app-lat-full}
\end{table*}

\begin{table*}[t]
\centering
\footnotesize
\begin{adjustbox}{max width=\textwidth}
\begin{tabular}{lrrrr}
\toprule
Workload (examples $\times$ repeats) & Query K1 & Risk K1 & Risk K2 & Hybrid K1 \\
\midrule
HotpotQA, original order (50 ex.\ $\times$ 25) & 0.109 / 0.161 & 1.129 / 1.553 & 1.135 / 1.558 & 1.267 / 1.745 \\
2Wiki, original order (50 ex.\ $\times$ 25) & 0.111 / 0.190 & 1.110 / 2.030 & 1.163 / 2.091 & 1.225 / 2.240 \\
2Wiki, score-neutral shuffled (50 ex.\ $\times$ 25) & 0.104 / 0.171 & 1.103 / 2.049 & 1.107 / 2.074 & 1.220 / 2.253 \\
$N{=}16$ nested, $p{=}1$ (50 ex.\ $\times$ 25) & 0.108 / 0.113 & 0.912 / 0.934 & 0.918 / 0.954 & 1.039 / 1.069 \\
Noise64 (67 documents) (50 ex.\ $\times$ 25) & 0.424 / 0.462 & 3.306 / 3.629 & 3.343 / 3.688 & 3.788 / 4.124 \\
\midrule
HotpotQA-500, gold-first (500 $\times$ 20) & 0.121 / 0.181 & 1.170 / 1.782 & 1.184 / 1.800 & --- \\
2Wiki-500, gold-first (500 $\times$ 20) & 0.105 / 0.197 & 1.041 / 2.079 & 1.053 / 2.125 & --- \\
Noise64, 67 docs (40 $\times$ 20) & 0.424 / 0.633 & 3.394 / 3.798 & 3.420 / 3.631 & --- \\
\bottomrule
\end{tabular}
\end{adjustbox}
\caption{CPU cost of the paper's selectors in milliseconds (median / p95). The timed region is feature computation, weighted scoring and top-$K$ selection for one selector call on the host CPU (x86\_64; CPU model not recorded); the suffix start is a single minimum over the selected positions and is timed only inside the end-to-end measurements of Table~\ref{tab:app-lat-full}, which include it. Upper block: inside the five L40S latency runs, median and p95 over the 50 per-example medians of 25 repeats. Lower block: the separate selector-only audit on the same host, median and p95 over all calls (Hybrid K1 was not timed there).}
\label{tab:app-plan}
\end{table*}

\begin{table*}[t]
\centering
\footnotesize
\begin{tabular}{lrrrr}
\toprule
Path & Ans. eq. [95\% CI] & Token exact & Gold EM & Changed-wrong \\
\midrule
Full prefill & 1.000 & 1.000 & 0.920 & 0 \\
Donor-copy Risk K1 (diagnostic) & 0.640 [0.50, 0.76] & 0.640 & 0.600 & 18 \\
Actual-forwarding Risk K1 & 0.440 [0.30, 0.58] & 0.440 & 0.440 & 27 \\
\bottomrule
\end{tabular}
\caption{Noise64 quality gate on the exact 50 examples and suffixes timed in Table~\ref{tab:latency-main} (mean repair ratio 0.519 for both paths; identical selected documents). Donor-copy and actual forwarding differ by +0.20 [0.06, 0.34] in normalized answer preservation (paired bootstrap). The mechanical median speed-up of Risk K1 on this workload therefore does not correspond to a behavior-preserving operating point.}
\label{tab:app-noise64-gate}
\end{table*}

\begin{figure*}[t]
\centering
\includegraphics[width=0.98\linewidth]{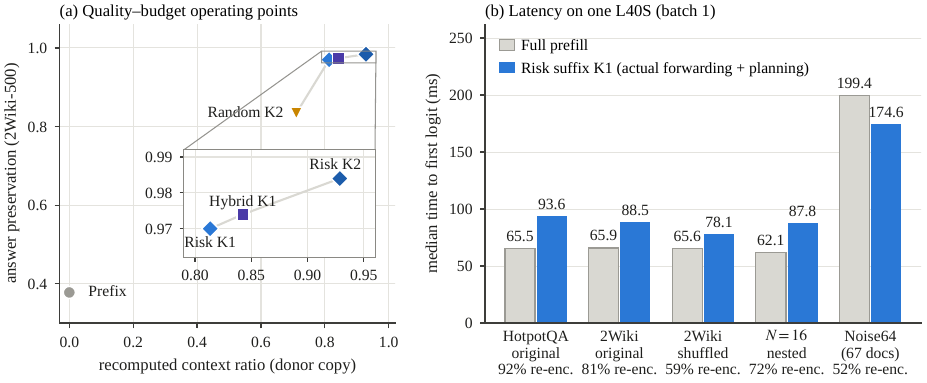}
\caption{Quality--budget and latency view. Left: 2Wiki-500 operating points under the donor-copy protocol (values of Table~\ref{tab:external-main}; inset: high-fidelity cluster). Right: median time to first logit for full prefill and risk suffix K1 with actual forwarding and selector planning on one L40S (values of Table~\ref{tab:latency-main}; percentages are the mean re-encoded context fraction). Risk K1 is slower on four workloads, and the Noise64 median crossover preserves only 44\% of full-prefill answers with worse p95 latency, so the operating rule serves every workload by full prefill. Every plotted value is listed in \texttt{figures/figure4\_budget\allowbreak\_gate\allowbreak\_pareto.data.csv}.}
\label{fig:budget-latency}
\end{figure*}

\paragraph{What the latency study captures.}
Warm, repeated, batch-1 time to first logit for one request with one composed cache resident on the GPU. The SAFE-KV timings include CPU selector scoring, suffix-start planning, the actual suffix forward, and the query forward; they exclude one-shot composed-cache construction (reported separately, 0.2--2.2\,s per example), cache lookup and transfer, admission and eviction, batching and scheduling, token emission, and any serving-stack effect. Every timed call returns a generation-ready cache whose first logits match an untimed reference bitwise. Paired statistics use per-example repeat medians and a 5{,}000-replicate example bootstrap.

\paragraph{Reading the results.}
On the four shorter or high-repair workloads (Figure~\ref{fig:budget-latency}, right), every SAFE-KV operating point is slower than full prefill at the paired median, and the descriptive p95 is worse for all of them. On Noise64, risk K1 and query K1 show median crossovers (1.149 and 1.120), but the query-K1 interval includes no improvement, both p95 values are worse than full prefill, and, on the same 50 examples, actual forwarding of the risk-K1 suffix preserves only 44\% of full-prefill answers (Table~\ref{tab:app-noise64-gate}). The prefix-query row evaluates the query against the unrepaired cache; it is fast because it repairs nothing and serves only to separate query evaluation from repair work. The crossover grows with context length, but at the repair budgets our selectors choose it does not pass the answer-preservation condition.

\paragraph{Selector cost.}
Table~\ref{tab:app-plan} reports the CPU cost of the paper's selectors. The timed region is feature computation, weighted scoring, and top-$K$ selection for one call on the host CPU (model not recorded); the suffix start is a single minimum over the selected positions and is timed only inside the end-to-end measurements, which include it. The selectors read no gold label, gold answer, full-prefill state, or mismatch mask. Selector overhead is thus small next to the cost of re-encoding the suffix, which takes most of the time of every SAFE-KV call in Table~\ref{tab:app-lat-full}.

\paragraph{Toward a deployment study.}
A deployment study would add serving-stack integration (e.g., vLLM), continuous batching and concurrency, a cache storage hierarchy and CPU--GPU transfer policy, prefetch scheduling, kernel-level cache-layout optimization, and production request traces, and would report TTFT, throughput, and memory under concurrent load.

\section{Representative Failure Patterns}

The result directories include per-run \texttt{answer\_failure\_analysis.md} files. We summarize representative patterns rather than relying on isolated anecdotes.

\paragraph{External HotpotQA drift.}
In the Qwen2.5-7B HotpotQA-500 run, prefix-conditioned composition changes 264 of 500 answers relative to full prefill, and 241 changed answers are also gold-wrong. High-KL examples include cases where full prefill returns the gold entity while prefix composition emits an unrelated entity or an ``information not provided'' style answer. Risk suffix K1 leaves only two changed-wrong answers and risk suffix K2 leaves one.

\paragraph{External 2Wiki drift.}
The 2Wiki run is harder. Prefix composition preserves only 37.8\% of full-prefill answers and produces 272 changed-wrong answers. Many high-KL failures are relation-chain questions where the composed cache loses which entity connects two documents. Risk K2 reduces changed-wrong errors to six, while random K2 leaves 66, illustrating why selected causal starts matter even though random suffix recomputation is strong.

\paragraph{Position-driven failures.}
The position sweep shows a clean causal pattern. When the answer-bearing document is first, prefix composition often answers that the information is unavailable even though the evidence is present. Oracle document patching alone does not help in this first-position condition because later distractor documents still contain stale states. Suffix repair recomputes those downstream states and recovers the full-prefill answer.

\paragraph{Selector misses.}
Remaining SAFE-KV failures fall into two groups. First, the selector can start too late, leaving an earlier causal dependency stale. Second, full prefill itself may produce a verbose or partially wrong answer, which makes exact answer equivalence and gold F1 diverge. These cases motivate better start-point prediction and semantic equivalence metrics beyond exact strings.

\section{Extended Causal-Closure Analysis and Localized Baselines}
\label{sec:app-closure}

Repairing the evidence document is not enough, and Table~\ref{tab:app-noise-closure} shows why under the donor-copy protocol. In low-noise diagnostics, oracle evidence-document patching looks strong because few later states are affected by distractors. Under high-overlap noise the same operation collapses: on Noise16-250 it reaches only 54.4\% answer preservation despite using oracle evidence labels, and random or query-selected document patching with larger $K$ also remains far below suffix repair.

\begin{table*}[t]
\centering
\footnotesize
\begin{adjustbox}{max width=\textwidth}
\begin{tabular}{llrrrr}
\toprule
Setting & Method & Ans. eq. & Gold EM & Changed-wrong & Patch \\
\midrule
Noise0-250 & Prefix-comp. & 0.672 & 0.760 & 60 & 0.000 \\
Noise0-250 & Oracle doc patch & 0.932 & 0.884 & 16 & 0.456 \\
Noise0-250 & Random doc K2 & 0.860 & 0.828 & 29 & 0.529 \\
\midrule
Noise8-250 & Prefix-comp. & 0.564 & 0.592 & 102 & 0.000 \\
Noise8-250 & Oracle doc patch & 0.724 & 0.772 & 56 & 0.193 \\
Noise8-250 & Random doc K2 & 0.644 & 0.676 & 81 & 0.164 \\
\midrule
Noise16-250 & Prefix-comp. & 0.416 & 0.420 & 145 & 0.000 \\
Noise16-250 & Oracle doc patch & 0.544 & 0.552 & 111 & 0.122 \\
Noise16-250 & Random doc K4 & 0.580 & 0.592 & 102 & 0.200 \\
Noise16-250 & Oracle suffix closure & 0.936 & 0.896 & 10 & 0.654 \\
Noise16-250 & All-doc patch sanity & 0.996 & 0.904 & 1 & 0.969 \\
\bottomrule
\end{tabular}
\end{adjustbox}
\caption{Noise-sweep and closure diagnostics (Qwen2.5-7B-Instruct, synthetic sets, donor-copy protocol). As high-overlap distractors increase, document-only patching degrades sharply even with oracle evidence labels; among the units tested, oracle suffix repair is the only one that approaches full-prefill behavior short of re-encoding every document.}
\label{tab:app-noise-closure}
\end{table*}

If the problem were simply locating the evidence text, oracle document patching would solve it; it does not. The missing object is the downstream state: later documents and the query are conditioned on histories that omit the evidence representation. Suffix repair re-encodes every later document state that could depend on the selected start. Linked-interval variants that re-encode a plausible local chain improve over document-only patching but remain below suffix repair on Noise16 (80.0\% for oracle linked intervals versus 93.6\% for oracle suffix repair), so human-interpretable reasoning chains are not guaranteed to align with the cache dependencies. All-document patching nearly recovers full prefill (99.6\%) and validates the patch operator, but spends almost the entire budget.

\paragraph{Operational query-attention token-local baseline.}
Table~\ref{tab:app-local-full} gives the full budget curves behind Table~\ref{tab:local-operational}. At every budget and in every order, attention-ranked spans preserve more answers than equal-budget random spans, and the gap widens with budget (at 60\%: 85/79/84\% versus 59/61/54\% mean over seeds for natural/shuffled/BM25). Below the 60\% budget the baseline still leaves substantial drift, and it does not reach the high-fidelity suffix operating points of Table~\ref{tab:actual-forwarding}; its planning also needs a separate query forward over the stale cache. The mean first-evidence rank of these conversions is 5.39 (natural), 5.42 (shuffled), and 3.23 (BM25) out of about ten paragraphs.

\begin{table*}[t]
\centering
\footnotesize
\begin{adjustbox}{max width=\textwidth}
\begin{tabular}{lrrrrrrr}
\toprule
Method / document-token budget & 0.05 & 0.1 & 0.16 & 0.2 & 0.3 & 0.45 & 0.6 \\
\midrule
\multicolumn{8}{l}{\emph{natural order} (prefix-composition 0.330; all-fresh chunked reference 1.000)} \\
Query-attention local & 0.500 & 0.470 & 0.540 & 0.590 & 0.620 & 0.720 & 0.850 \\
\quad token exact & 0.490 & 0.460 & 0.520 & 0.590 & 0.620 & 0.720 & 0.840 \\
Random local, seed 13 & 0.330 & 0.410 & 0.380 & 0.370 & 0.480 & 0.550 & 0.650 \\
Random local, seed 29 & 0.340 & 0.360 & 0.350 & 0.440 & 0.450 & 0.540 & 0.580 \\
Random local, seed 47 & 0.350 & 0.360 & 0.360 & 0.420 & 0.420 & 0.500 & 0.540 \\
Context ratio & 0.049 & 0.098 & 0.157 & 0.196 & 0.294 & 0.441 & 0.589 \\
\midrule
\multicolumn{8}{l}{\emph{shuffled order} (prefix-composition 0.430; all-fresh chunked reference 0.990)} \\
Query-attention local & 0.470 & 0.500 & 0.580 & 0.580 & 0.670 & 0.740 & 0.790 \\
\quad token exact & 0.440 & 0.500 & 0.560 & 0.560 & 0.650 & 0.730 & 0.780 \\
Random local, seed 13 & 0.420 & 0.430 & 0.390 & 0.450 & 0.450 & 0.480 & 0.630 \\
Random local, seed 29 & 0.420 & 0.420 & 0.390 & 0.460 & 0.470 & 0.550 & 0.650 \\
Random local, seed 47 & 0.420 & 0.420 & 0.450 & 0.440 & 0.450 & 0.580 & 0.540 \\
Context ratio & 0.049 & 0.098 & 0.157 & 0.196 & 0.294 & 0.441 & 0.589 \\
\midrule
\multicolumn{8}{l}{\emph{BM25 order} (prefix-composition 0.300; all-fresh chunked reference 1.000)} \\
Query-attention local & 0.370 & 0.450 & 0.480 & 0.500 & 0.640 & 0.730 & 0.840 \\
\quad token exact & 0.370 & 0.450 & 0.470 & 0.490 & 0.640 & 0.730 & 0.840 \\
Random local, seed 13 & 0.350 & 0.360 & 0.370 & 0.380 & 0.420 & 0.470 & 0.560 \\
Random local, seed 29 & 0.320 & 0.330 & 0.410 & 0.380 & 0.460 & 0.530 & 0.530 \\
Random local, seed 47 & 0.350 & 0.340 & 0.360 & 0.380 & 0.440 & 0.500 & 0.520 \\
Context ratio & 0.049 & 0.098 & 0.157 & 0.196 & 0.294 & 0.441 & 0.589 \\
\bottomrule
\end{tabular}
\end{adjustbox}
\caption{Operational query-attention token-local baseline: full budget curves on HotpotQA (100 examples per order), actual forwarding, Qwen2.5-7B-Instruct, one RTX PRO 6000 Blackwell GPU. Values are normalized answer preservation relative to the all-fresh chunked reference (token-sequence exact shown for the query-attention rows); Context ratio is the recomputed context-token fraction at each document-token budget. Random local uses equal-budget random spans with seeds 13/29/47.}
\label{tab:app-local-full}
\end{table*}

\paragraph{Full-reference token-local diagnostic.}
Table~\ref{tab:app-oracle-local} reports a deliberately optimistic diagnostic on 2Wiki in score-neutral shuffled order: token positions are ranked by the drift between the composed and full-prefill K/V states, and the selected states are donor-copied from the full-prefill cache. Because both steps read the reference, it is an analysis tool, not a deployable method. It shows that where tokens are placed matters, since drift-ranked positions beat equal-budget random positions at every budget, and at the risk-K1-matched context cost it exceeds actual risk-K1 suffix repair by 10 points [1, 20] in normalized answer preservation (9 points [0, 19] in generated-token exactness). Fifteen of the 100 examples lose normalized exactness as the nested oracle budget grows, so the diagnostic is not a monotone upper bound. Informed token placement is thus a promising route to cheaper repair.

\begin{table*}[t]
\centering
\footnotesize
\begin{adjustbox}{max width=\textwidth}
\begin{tabular}{lrrrrr}
\toprule
Method / document-token budget & 0.20 & 0.30 & 0.45 & 0.60 & Risk-K1-matched \\
\midrule
Prefix-comp. & 0.390 & 0.390 & 0.390 & 0.390 & --- \\
Oracle KV-drift token-local (donor copy) & 0.630 & 0.620 & 0.700 & 0.830 & 0.820 \\
Random positions, oracle full-KV donor & 0.450 & 0.490 & 0.570 & 0.630 & 0.690 \\
Risk suffix K1, actual forwarding & --- & --- & --- & --- & 0.720 \\
Risk suffix K1, donor copy & --- & --- & --- & --- & 0.770 \\
Context ratio & 0.192 & 0.288 & 0.432 & 0.576 & 0.622 \\
\bottomrule
\end{tabular}
\end{adjustbox}
\caption{Full-reference-informed token-local diagnostic on 2Wiki, score-neutral shuffled order (seed 101), 100 examples. Token positions are ranked by the K/V drift between the composed cache and the full-prefill cache, and the selected K/V states are donor-copied from the full-prefill cache; both steps use the reference, so this diagnostic is an analysis tool rather than a deployable method. Values are normalized answer preservation. At the Risk-K1-matched context cost the oracle-local minus actual-Risk-K1 paired difference is +0.10 [0.01, 0.20] (generated-token exact: +0.09 [0.00, 0.19]), a descriptive interval without multiple-comparison correction.}
\label{tab:app-oracle-local}
\end{table*}

\section{Selector Specification, Diagnostics, and Negative Results}
\label{sec:app-selector}

SAFE-KV separates the repair mechanism from the selector. The mechanism is suffix repair from a selected start; the selector only estimates that start. This appendix gives the executable specification of the shipped selectors, maps each term of Equations~\ref{eq:query} and~\ref{eq:risk} to the implementation, notes an unused configuration block, quantifies the tie-break, and reports the selector diagnostics.

\paragraph{Executable specification.}
All functions are in \texttt{code/safekv/doc\_selector.py} unless noted.
{\raggedright
\begin{itemize}
    \item \textbf{Terms} $T(x)$ (\texttt{lexical\_terms}, \texttt{patch\_recompute.py}): the set of lower-cased matches of \texttt{[A-Za-z0-9]+} in $x$, excluding the 21 stopwords \emph{a, an, and, at, by, does, for, from, in, is, of, on, or, the, to, was, what, when, where, who, with}.
    \item \textbf{Query score} $Q_i$ (\texttt{query\_similarity\_doc\_scores}, \texttt{lexical\_overlap}): $|T(q)\cap T(t_i)|/|T(q)|$ with $t_i$ the title and text of $d_i$ joined by a space; $0$ if $T(q)=\emptyset$.
    \item \textbf{Entities} $\mathrm{Ent}(x)$ (\texttt{extract\_entities}): lower-cased matches of (i) two or more consecutive capitalized words, (ii) all-capital tokens of length $\geq 2$ with optional \texttt{-}/\texttt{\_} suffixes, (iii) tokens containing a digit with optional \texttt{:}, \texttt{.}, \texttt{-} and AM/PM suffix, and (iv) single capitalized words not in \{a, an, and, archive, document, note, question, the\}.
    \item \textbf{Entity recall} $E_i$ (\texttt{entity\_overlap\_doc\_scores}): $|\mathrm{Ent}(q)\cap\mathrm{Ent}(t_i)|/|\mathrm{Ent}(q)|$; $0$ if $\mathrm{Ent}(q)=\emptyset$.
    \item \textbf{Conflict indicator} $I^{\mathrm{conf}}_i$ (\texttt{conflict\_candidate\_docs}, \texttt{answer\_candidate\_terms}, \texttt{has\_conflict\_like\_cue}): a document is a candidate if its text contains a number/time-like or month-day answer candidate and either the question contains one of \emph{according to, official, latest, current, schedule, trusted} or $Q_i\geq 0.25$; if the candidate value sets of the candidate documents are not all identical, every candidate receives $I^{\mathrm{conf}}_i=1$, otherwise all receive $0$.
    \item \textbf{Shared entities} $\hat X_i$ (\texttt{cross\_doc\_shared\_entity\_counts}, \texttt{normalize\_by\_doc\_id}): the number of entities of $t_i$ that occur in at least one other document, min--max normalized within the example.
    \item \textbf{Retrieval} $\hat r_i$ (\texttt{normalize\_by\_doc\_id}): the file's \texttt{retrieval\_score}, min--max normalized within the example; if all scores are equal, every document receives $1$ when they are positive and $0$ when they are zero.
    \item \textbf{Risk score} $R_i$ (\texttt{risk\_guided\_doc\_scores}): $0.25\hat r_i+0.30Q_i+0.20E_i+0.15I^{\mathrm{conf}}_i+0.10\hat X_i$.
    \item \textbf{Top-$K$} (\texttt{select\_top\_k\_doc\_ids}): sort by (score descending, document id ascending), take the first $K$.
    \item \textbf{Hybrid} (\path{run_answer_eval.py::_hybrid_query_risk_union_suffix_doc_ids}): $S=\{\arg\max Q\}\cup\{\arg\max R\}$.
    \item \textbf{Start} (\path{_suffix_closure_ids_from_selected}): the earliest prompt position among $S$; the suffix is every document from there to the end.
\end{itemize}
\par}

\paragraph{Implementation note.}
The shipped configuration files contain a legacy \texttt{risk} block with \texttt{similarity\_backend: tfidf} and weights (query 0.40, entity 0.20, retrieval 0.15, boundary 0.10, cross-document 0.10, conflict 0.05). The scorer does not read this block: \texttt{risk\_guided\_doc\_scores} takes no configuration argument, uses lexical recall rather than TF--IDF, has no boundary feature, and hard-codes the weights of Equation~\ref{eq:risk}. Every reported run executed Equations~\ref{eq:query} and~\ref{eq:risk} as written; the artifact README notes the unused block.

\paragraph{Inputs.}
The selectors read the question, document titles and texts, document order, and retrieval scores. They read no gold evidence label, gold answer, full-prefill state, or mismatch mask. In the gold-evidence-first construction and in the synthetic diagnostics the retrieval score is construction-assigned and gold-correlated (Appendix~\ref{sec:app-data}); the score-neutral shuffled runs set it to zero.

\paragraph{Tie-break audit.}
Table~\ref{tab:app-tie-break} measures how often the ascending-id tie-break decides the suffix start. Ties are the rule for the query score whenever several documents share the question terms: in the position sweep 18.7 of 19 documents tie on average and the tie-break selects the evidence document (id 0) in all 250 cases, so the ``query suffix K1'' row of Table~\ref{tab:position} equals the oracle-suffix row; in the Mistral noisy diagnostic 197 of 250 query-K1 selections are ties resolved to an evidence document; on the nested staleness inputs every query-K1 selection is a tie resolved to the earliest position (document 1 in 43 of 50 examples at $N{=}16$), and risk K1/K2 are tie-decided in 35 and 28 of 50 examples there because retrieval scores are zero. On the external conversions ids equal positions, so a tie always selects the earlier document, i.e., a longer suffix; it decides the start for risk K1 in 7 and 6 of 500 examples (HotpotQA/2Wiki), for risk K2 in 1 and 0, for hybrid K1 in 38 and 50, and for query K1 and K2 in 183--191 and 183 and 174 of 500, usually toward an evidence document because evidence comes first; in the 2Wiki-100 subset of Table~\ref{tab:local-vs-suffix} risk K1 is tie-decided in 1 of 100 examples. The headline risk rows of Tables~\ref{tab:external-main} and~\ref{tab:actual-forwarding}(a), the local-versus-suffix comparison of Table~\ref{tab:local-vs-suffix}, and the suffix-repair conclusion are therefore essentially unaffected, whereas the query rows on the gold-informed and synthetic constructions are partly or wholly tie-driven, which is why selector conclusions rest on the risk rows and the score-neutral variants, where the tie is position-only. Elsewhere the risk selector is tie-decided in at most 7\% of examples.

\begin{table*}[t]
\centering
\footnotesize
\setlength{\tabcolsep}{4pt}
\begin{adjustbox}{max width=\textwidth}
\begin{tabular}{llrrrrrr}
\toprule
Data & id $=$ pos. & Query K1 & Query K2 & Hybrid K1 & Risk K1 & Risk K2 & Q-K1 tie $\to$ evid. \\
\midrule
HotpotQA-500, gold-first & yes & 183/500 & 183/500 & 38/500 & 7/500 & 1/500 & 125/183 \\
2Wiki-500, gold-first & yes & 191/500 & 174/500 & 50/500 & 6/500 & 0/500 & 179/191 \\
2Wiki-100, gold-first (Table 3) & yes & 34/100 & 39/100 & 10/100 & 1/100 & 0/100 & 30/34 \\
HotpotQA-100, score-neutral shuffle 101 & yes & 40/100 & 52/100 & 26/100 & 7/100 & 2/100 & 12/40 \\
2Wiki-100, score-neutral shuffle 101 & yes & 34/100 & 44/100 & 21/100 & 5/100 & 3/100 & 17/34 \\
Position sweep (Noise16, 19 docs) & no & 250/250 & 250/250 & 173/250 & 0/250 & 16/250 & 250/250 \\
Noise16-250 & no & 150/250 & 213/250 & 95/250 & 2/250 & 3/250 & 50/150 \\
Noisy diagnostic (Mistral rows, first 250) & no & 197/250 & 91/250 & 106/250 & 0/250 & 3/250 & 197/197 \\
Noise64-50 & no & 30/50 & 41/50 & 20/50 & 1/50 & 1/50 & 10/30 \\
Nested $N{=}16$, layout 101 & yes & 50/50 & 50/50 & 50/50 & 35/50 & 28/50 & 1/50 \\
\bottomrule
\end{tabular}
\end{adjustbox}
\caption{Tie-break audit of the shipped selectors. For each dataset and selector, the middle columns count examples whose suffix-recompute start would change if score ties were broken by descending instead of ascending document id, i.e., examples where the id tie-break decides the start (for hybrid K1, through its query top-1 component). Wherever ids equal positions, the ascending rule always picks the earlier start, i.e., a longer suffix. ``id $=$ pos.'' states whether document ids equal prompt positions in that file. The last column (``Q-K1 tie $\to$ evid.'') counts, among Query-K1 examples with a top-1 tie, how many the ascending-id rule resolves to a gold evidence document. In the synthetic generator ids are assigned before shuffling and evidence documents receive the smallest ids, so a tie there favors evidence; in the external conversions ids equal prompt positions and evidence documents come first, so a tie favors evidence there too; only the score-neutral shuffled variants make the tie-break position-only. The risk selector is rarely tie-decided when retrieval scores are present; on the zero-score nested inputs it is tie-decided in 35 and 28 of 50 examples.}
\label{tab:app-tie-break}
\end{table*}

\begin{table*}[t]
\centering
\footnotesize
\begin{adjustbox}{max width=\textwidth}
\begin{tabular}{llrrrrr}
\toprule
Setting & Method & Ans. eq. & Gold EM & Gold F1 & Changed-wrong & Patch \\
\midrule
2Wiki-500 & Risk suffix K1 & 0.970 & 0.442 & 0.536 & 12 & 0.813 \\
2Wiki-500 & Margin-adaptive & 0.972 & 0.440 & 0.534 & 12 & 0.838 \\
2Wiki-500 & Hybrid suffix K1 & 0.974 & 0.442 & 0.536 & 11 & 0.842 \\
2Wiki-500 & Risk suffix K2 & 0.984 & 0.452 & 0.546 & 6 & 0.929 \\
\midrule
Mistral noisy-250 & Query suffix K1 & 0.908 & 0.392 & 0.616 & 18 & 0.679 \\
Mistral noisy-250 & Risk suffix K1 & 0.852 & 0.376 & 0.607 & 31 & 0.668 \\
Mistral noisy-250 & Hybrid suffix K1 & 0.928 & 0.388 & 0.613 & 14 & 0.790 \\
Mistral noisy-250 & Query suffix K2 & 0.996 & 0.392 & 0.612 & 1 & 0.872 \\
Mistral noisy-250 & Risk suffix K2 & 0.996 & 0.392 & 0.612 & 1 & 0.879 \\
\bottomrule
\end{tabular}
\end{adjustbox}
\caption{Selector and budget operating points (donor-copy protocol). 2Wiki-500 rows use the gold-evidence-first construction; Mistral rows use the synthetic noisy diagnostic, where evidence documents carry the smallest ids and the query and hybrid rows are largely decided by the ascending-id tie-break toward evidence (Table~\ref{tab:app-tie-break}), and random suffix K2 reaches 1.000 on that set. Hybrid K1 is a balanced refinement when the K1 selectors disagree; K2 remains the high-fidelity setting.}
\label{tab:selector-budget}
\end{table*}

\begin{table*}[t]
\centering
\footnotesize
\begin{adjustbox}{max width=\textwidth}
\begin{tabular}{llrrrrr}
\toprule
Data & Selector & Recall & Precision & Any-hit & Exact-set & Avg suffix \\
\midrule
HotpotQA-500 & Query K2 & 0.544 & 0.544 & 0.848 & 0.240 & 0.886 \\
HotpotQA-500 & Risk K2 & 0.858 & 0.858 & 1.000 & 0.716 & 0.973 \\
HotpotQA-500 & Learned K2 & 0.008 & 0.008 & 0.012 & 0.004 & 0.504 \\
HotpotQA-500 & Random K2 & 0.429 & 0.429 & 0.752 & 0.106 & 0.826 \\
\midrule
2Wiki-500 & Query K2 & 0.657 & 0.740 & 0.972 & 0.376 & 0.934 \\
2Wiki-500 & Risk K2 & 0.818 & 0.920 & 0.998 & 0.638 & 0.969 \\
2Wiki-500 & Learned K2 & 0.000 & 0.000 & 0.000 & 0.000 & 0.490 \\
2Wiki-500 & Random K2 & 0.412 & 0.470 & 0.788 & 0.098 & 0.809 \\
\bottomrule
\end{tabular}
\end{adjustbox}
\caption{Selector evidence-recall diagnostics on the gold-evidence-first external conversions (retrieval scores construction-assigned; Appendix~\ref{sec:app-data}). The synthetic-trained learned selector does not transfer to external RAG evidence labels; the high any-hit and precision of risk-guided selection here reflect the gold-correlated retrieval score (compare Table~\ref{tab:actual-forwarding}(b)). Avg suffix is the average fraction of document positions re-encoded after suffix repair.}
\label{tab:app-selector-diagnostics}
\end{table*}

\paragraph{Evidence recall on the gold-informed construction.}
Table~\ref{tab:app-selector-diagnostics} reports evidence-recall diagnostics on the external conversions. Risk K2 has any-hit rates of 100.0\% on HotpotQA and 99.8\% on 2Wiki, which explains the high preservation of Table~\ref{tab:external-main} and its cost: the selected start is often early. These rates reflect the gold-informed retrieval score and evidence-first order. On the score-neutral shuffled inputs, a selector-only audit finds risk-K2 evidence recall of 0.467 (HotpotQA) and 0.620 (2Wiki) and exact-set rates of 0.163 and 0.321, consistent with the lower preservation of the risk selectors in Table~\ref{tab:actual-forwarding}(b).

\paragraph{Query selection.}
Query K2 has lower evidence recall than risk K2 on the gold-informed construction but is the more reliable operating point once the construction cues are removed (Table~\ref{tab:actual-forwarding}(b)), at about 86\% recomputation. Its behavior on synthetic data is dominated by the tie-break described above.

\paragraph{Synthetic learned selection did not transfer.}
A learned selector trained on synthetic noise has strong synthetic dev behavior but near-zero evidence recall on the external conversions (Table~\ref{tab:app-selector-diagnostics}). This is why SAFE-KV uses training-free heuristics as operating policies and places the weight of the method on suffix repair.

\paragraph{Random controls are strong by design.}
Once a random selector starts early enough, suffix repair re-encodes most of the downstream state, which is why the main text reports paired comparisons against random K2 rather than easier baselines; on the Mistral diagnostic, where random K2 reaches 100\%, the mechanism rather than the selector carries the result.

\section{Design Ledger: How the Experiments Shaped the Method}
\label{sec:app-design-ledger}

Table~\ref{tab:design-ledger} records the experimental path behind the SAFE-KV design. The method is not the best row of one benchmark table but the conclusion of a sequence of positive and negative results: non-prefix composition creates behavioral drift; local evidence repair is insufficient on the tested diagnostics; linked or fixed local intervals remain brittle; full repair is safe but expensive; and selected suffix repair is the smallest unit that consistently recovered full-prefill behavior. All rows use the donor-copy protocol and the constructions of Appendix~\ref{sec:app-data}.

\begin{table*}[t]
\centering
\footnotesize
\begin{tabular}{@{}>{\raggedright\arraybackslash}p{0.13\textwidth}>{\raggedright\arraybackslash}p{0.29\textwidth}>{\raggedright\arraybackslash}p{0.28\textwidth}>{\raggedright\arraybackslash}p{0.21\textwidth}@{}}
\toprule
Experiment & Key result & Interpretation & Design decision \\
\midrule
External 7B composition & Prefix composition preserves only 47.2\% HotpotQA and 37.8\% 2Wiki full-prefill answers. & Non-prefix document reuse is a behavioral problem even after matching the instruction prefix. & Make full-prefill behavior preservation the primary contract. \\
Document-only repair & On Noise16, oracle evidence-document patching reaches only 54.4\% equivalence. & Gold evidence labels are not causally closed cache repair units. & Repair causally closed units rather than evidence documents. \\
Downstream windows & On 2Wiki-100, w4 reaches 79.0\% equivalence at 72.9\% patch ratio; suffix K1 reaches 97.0\% at 79.3\%. & Repair quality changes sharply when the recomputed region crosses the causal frontier. & Use suffix closure rather than a fixed local window. \\
Linked intervals & On Noise16, oracle linked intervals reach 80.0\%, below oracle suffix closure at 93.6\%. & Human-plausible reasoning chains do not always match transformer cache dependencies. & Keep linked intervals as a negative ablation. \\
Adaptive suffix & On Noise16, risk adaptive suffix keeps 86.8\% equivalence at 57.9\% patch ratio versus risk suffix K2 at 88.8\% and 69.9\%. & Budget can be traded for fidelity, but aggressive trimming increases residual drift. & Present SAFE-KV as a quality--budget family. \\
Learned selector & On external selector diagnostics, synthetic-trained learned selection has near-zero evidence recall. & Selector learning is sensitive to supervision shift; the stable contribution is the repair contract. & Use lightweight query/risk selectors in the main paper and report the learned-selector failure. \\
Random suffix controls & Random K2 reaches 91.8\% HotpotQA and 83.4\% 2Wiki, but selected risk K2 reaches 99.8\% and 98.4\%. & Coherent suffix recomputation is itself powerful, so the control is intentionally strong. & Report paired improvements over random rather than relying on easier controls. \\
All-document sanity & All-document patching reaches 100.0\% equivalence on noisy sanity cases at 89.9\% patch ratio. & The patch operator can recover the oracle when enough causal state is recomputed. & Use all-document patching only as an upper-bound sanity check. \\
\bottomrule
\end{tabular}
\caption{Design ledger for SAFE-KV (donor-copy protocol). Each row records an experimental route that shaped the algorithm; suffix repair is the simplest repair unit that remained consistent across these positive and negative results.}
\label{tab:design-ledger}
\end{table*}

\paragraph{Why this matters for interpretation.}
The negative results rule out simpler stories. If oracle evidence-document repair had solved the high-noise diagnostics, SAFE-KV would be a document-selection method. If fixed downstream windows had matched suffix repair, it would be a local repair heuristic. If synthetic-trained learned selectors had transferred to external data, selector learning would be the main contribution. The data point elsewhere: the robust object is the suffix, and the selector is a budgeted estimate of where that suffix should begin. The later experiments map the operating regime: repair matters at larger document counts, the selectors need per-workload validation without construction cues, informed token-local repair recovers much but not all of the drift, and at today's budgets full prefill remains the faster faithful choice.

\paragraph{How this should guide future systems.}
A serving system can choose a more aggressive point on the quality--budget frontier, such as adaptive suffix repair, but it should expose that choice as a risk decision. The SAFE-KV protocol makes the decision auditable: compare against full prefill on the exact prompt, report changed-and-wrong answers, compare against matched random suffix controls, separate donor-copy diagnostics from actual forwarding, and report recomputed ratios and measured latency separately.

\section{Result Inventory and Compact Development Tables}
\label{sec:app-result-inventory}

This appendix records the role of every result family in the consolidated artifact. Counting every file under \path{results/} (main donor-copy runs, 323 files), \path{results_rebuttal/} (actual-forwarding, staleness, latency, and reconciliation runs and their input variants, 307 files), and \path{results_august/run_registry/} (token-local baseline runs, 18 files), the artifact contains \inventoryResultFiles{} result files, of which \inventoryAnswerFiles{} are answer-level metric files (\path{answer_eval_metrics.csv}; eight of them are small-model development mirrors under \path{results/metrics/}); eleven of the \inventoryResultFiles{} are the raw per-example dumps attached to the release rather than stored in git. The artifact ships a generated listing of these files with sizes, SHA-256 values and locations (\path{RESULT_FILE_LISTING.txt}) and a repository-wide \path{SHA256SUMS}. Runs play different roles: smoke runs validate code paths, small-model runs document development history, negative linked/adaptive/learned-selector runs constrain the method, and the actual-forwarding and latency runs map the operating regime. Table~\ref{tab:app-result-inventory} maps each family to the role it plays in the paper's tables, figures, and claims.

\begin{table*}[t]
\centering
\footnotesize
\begin{adjustbox}{max width=\textwidth}
\begin{tabular}{p{0.20\textwidth}p{0.18\textwidth}p{0.24\textwidth}p{0.30\textwidth}}
\toprule
Result family & Scale & Representative files & How it is used in the paper \\
\midrule
Main external 7B & HotpotQA-500 and 2Wiki-500, with bootstrap and held-out splits & Hotpot/2Wiki suffix-main runs & Headline behavior-preservation evidence, random-control comparisons, full-prefill-correct subsets, and calibration checks. \\
External development sweeps & 100--200 example HotpotQA/2Wiki sweeps & K, hybrid, margin, adaptive, and cost sweeps & Chose risk K1, hybrid K1, risk K2, and adaptive variants as quality--budget operating points rather than test-set-only choices. \\
High-noise diagnostics & Noise0/8/16, 250 examples per setting, five controlled categories & Noise sweep, closure, and K-sweep runs & Establish that document-only repair degrades with distractor load and that, among the units tested, suffix repair is the one that approaches full-prefill behavior. \\
Locality stress tests & Downstream windows, linked intervals, adaptive suffixes & Window, linked-interval, and adaptive runs & Negative evidence against fixed local windows and human-plausible linked intervals; motivates causal suffix closure. \\
Position sensitivity & 50 examples across five evidence positions & Position-sweep suffix runs & Lost-in-the-middle style test showing why early evidence is hardest for document-only patching and why suffix repair removes most position dependence. \\
Cross-model checks & Mistral-7B noisy 250 plus smoke runs & Mistral noisy suffix runs & Shows the composition-drift and suffix-repair pattern beyond Qwen2.5-7B. \\
Prototype latency and memory & 2Wiki short prompts and longer noise64 prompts, donor copy, no planning & 2Wiki and noise64 prototype latency runs & Early timing scope (Table~\ref{tab:app-cost}); the L40S actual-forwarding study is used instead. \\
Actual forwarding & HotpotQA/2Wiki-500 original; 3$\times$100 score-neutral shuffled; same-run donor/actual reconciliation & \path{actual_recompute_*}, \path{neutral_selector_actual_*}, \path{actual_vs_copy_*} & Deployable-protocol evidence for the main operating points and the selector scope. \\
Staleness $\times$ document count & $5\times5$ grid, 50 paired examples, 3 mask seeds, 7 methods & \path{staleness_doccount_full_50_v3} & Where repair is needed as a function of $N$ and $p$. \\
L40S latency & five 50-example workloads, 5 repeats; Noise64 paired quality gate & \path{latency_*_50_v1}, \path{quality_noise64_actual_riskk1_50_v1} & Main-text latency table and operating rule. \\
Localized baselines & query-attention local, 3 orders $\times$ 100; oracle KV-drift local, 100 & \path{qwen25_query_attention_*}, \path{oracle_kv_drift_*} & Tested localized-repair baselines. \\
Development and sanity & Small-model tables, all-document patch, smoke runs, learned selector artifacts & Metrics, all-doc sanity, learned-selector, and smoke runs & Diagnostic evidence, implementation sanity, and the negative selector-transfer result. \\
\bottomrule
\end{tabular}
\end{adjustbox}
\caption{Run-level inventory for the consolidated artifact (\inventoryResultFiles{} result files, \inventoryAnswerFiles{} answer-level metric files; see the generated listing in the repository). The table records how every result family is used: headline evidence, mechanism analysis, negative ablation, calibration, latency, operating-regime analysis, or sanity/development support.}
\label{tab:app-result-inventory}
\end{table*}

\paragraph{How to read strong random baselines.}
Moving from prefix composition to random suffix repair measures the benefit of coherent downstream re-encoding; moving from random suffix repair to selected suffix repair measures whether the selector estimates the start better than chance. This is why random suffix is treated as a substantive control.

\paragraph{Compact development tables.}
Tables~\ref{tab:composition}, \ref{tab:granularity}, and~\ref{tab:answer-summary} are compact summaries of the small-model development runs. Their numbers are contained in the larger appendix tables; they are kept because they show the development path: prefix-conditioned composition was chosen as the strongest simple baseline; sub-document span and window units did not separate from matched random patching on first-token metrics (Table~\ref{tab:granularity} contains no document-suffix unit); and small-model answer-level diagnostics showed the same qualitative composition failure before the 7B experiments became the main evidence.

\begin{table}[t]
\centering
\footnotesize
\begin{adjustbox}{max width=\linewidth}
\begin{tabular}{cccc}
\toprule
Method & KL & Top-1 & Top-10 \\
\midrule
Doc-only & 0.854 & 0.61 & 0.56 \\
Prefix-cond. & 0.207 & 0.90 & 0.70 \\
Prefix-plus & 1.205 & 0.47 & 0.57 \\
\bottomrule
\end{tabular}
\end{adjustbox}
\caption{Early development diagnostic (Qwen2.5-0.5B-Instruct, clean diagnostic set, 100 examples): first-token KL and top-1/top-10 agreement with full prefill for three composition variants. Prefix-conditioned document composition is far safer than doc-only or prefix-plus composition but still not exact; it is the baseline used throughout the paper.}
\label{tab:composition}
\end{table}

\begin{table}[t]
\centering
\footnotesize
\begin{adjustbox}{max width=\linewidth}
\begin{tabular}{cccc}
\toprule
Repair unit & KL & Top-1 & Patch ratio \\
\midrule
Base & 0.207 & 0.90 & 0.00 \\
Span & 0.232 & 0.91 & 0.14 \\
Span+16 & 0.204 & 0.92 & 0.24 \\
S+C suffix & 0.180 & 0.96 & 0.22 \\
Matched random & 0.150 & 0.95 & 0.22 \\
Oracle doc & 0.090 & 0.97 & 0.46 \\
Full-doc sanity & 0.000 & 1.00 & 0.89 \\
\bottomrule
\end{tabular}
\end{adjustbox}
\caption{Early development diagnostic (Qwen2.5-0.5B-Instruct, clean diagnostic set, 100 examples, donor-copy protocol): first-token KL to full prefill, top-1 agreement, and patch ratio for sub-document repair units. ``Span'' re-encodes the evidence span, ``Span+16'' extends it by 16 tokens, and ``S+C suffix'' re-encodes from the support/conflict sentence to the end of its document; none of these is the document-suffix repair of Section~\ref{sec:problem}. On this small model the first-token metrics do not separate these units from matched random patching; the answer-level 7B results (Tables~\ref{tab:external-main}--\ref{tab:local-vs-suffix}) are the evidence for the paper's conclusions about repair units.}
\label{tab:granularity}
\end{table}

\begin{table*}[t]
\centering
\footnotesize
\begin{adjustbox}{max width=\textwidth}
\begin{tabular}{cccccccc}
\toprule
Model & Split & Base & Oracle & Query/Entity & Gate .60 & Random K2 & Wrong changes \\
\midrule
Qwen0.5B & clean & 0.85 & 0.97 & 0.97/0.97 & 0.92 & 0.96 & 14 \\
Qwen0.5B & noisy & 0.69 & 0.90 & 0.89/0.90 & 0.81 & 0.83 & 23 \\
Qwen1.5B & clean & 0.88 & 0.96 & 0.96/0.96 & 0.92 & 0.93 & 11 \\
Qwen1.5B & noisy & 0.74 & 0.89 & 0.92/0.92 & 0.84 & 0.86 & 20 \\
\bottomrule
\end{tabular}
\end{adjustbox}
\caption{Early small-model development runs (Qwen2.5-0.5B and 1.5B, clean and noisy diagnostic splits, 100 examples each, donor-copy protocol). Values are normalized answer preservation relative to full prefill, except the final column, which counts prefix-composed answers that changed from full prefill and were gold-wrong. On these models, replacing two selected documents (oracle, query, or entity K2) reduces changed answers relative to no repair and random K2 is nearly as strong; these development runs predate suffix repair and the 7B experiments.}
\label{tab:answer-summary}
\end{table*}

\section{Figure Provenance}
\label{sec:app-figures}

Every figure is drawn at the physical size at which it is included, with TrueType fonts and no text below 7.5\,pt. Every data figure is plotted from a CSV of its values shipped in the artifact under \path{paper_reference/final_figures/}, and \path{scripts/camera_ready/make_figures.py} regenerates these CSVs from the raw result files.
{\raggedright
\begin{itemize}
    \item Figure~\ref{fig:setup} is the conceptual contract (no data): full prefill is the reference state, non-prefix composition is a context-mismatched state, and suffix repair re-encodes the downstream documents from the selected start.
    \item Figures~\ref{fig:local-vs-suffix} and~\ref{fig:position-sweep} (main text) plot the downstream-window results of Table~\ref{tab:local-vs-suffix} and the position sweep of Table~\ref{tab:position}. Plotted values: \path{figures/figure_local_vs_suffix.data.csv} and \path{figures/figure_position_sweep.data.csv}.
    \item Figure~\ref{fig:app-robustness} plots the full-prefill-correct subsets of Table~\ref{tab:cross-external} (recomputed from the per-example rows) and the Mistral-7B diagnostic of Table~\ref{tab:selector-budget} (donor-copy protocol); plotted values in \path{figures/figure5_external_sanity.data.csv}.
    \item Figure~\ref{fig:budget-latency} plots Table~\ref{tab:external-main} (left) and Table~\ref{tab:latency-main} (right); plotted values in \path{figures/figure4_budget_gate_pareto.data.csv}.
\end{itemize}
\par}

\section{Extended Related-Work Map}
\label{sec:app-reference-map}

This appendix expands the main related-work discussion by making explicit how SAFE-KV sits at the intersection of four literatures.

\paragraph{Serving and cache systems.}
Exact-prefix and serving systems motivate the efficiency pressure behind SAFE-KV: paging, scheduling, disaggregation, cache storage, transfer, compression, and modular reuse, and selective recomputation of chunk caches \citep{kwon2023efficientmemorymanagementlarge,sheng2023flexgen,zhong2024distserve,agrawal2024sarathiserve,patel2023splitwise,zheng2024sglang,liu2025lmcacheefficientkvcache,liu2023cachegen,gim2024promptcachemodularattention,yao2025cacheblendfastlargelanguage,agarwal2025cachecraft,lu2024turborag,yang2025kvlinkacceleratinglargelanguage}. SAFE-KV contributes a semantic contract for the non-prefix regime these systems increasingly approach.

\paragraph{KV memory, long context, and position.}
Compression and long-context work show that KV state is a first-class computational object \citep{dao2022flashattention,dao2023flashattention2,shazeer2019fasttransformerdecoding,ainslie2023gqa,liu2024kivituningfreeasymmetric2bit,liu2023scissorhandsexploitingpersistenceimportance,zhang2023h2o,xiao2023streamingllm,li2024snapkv,ge2023fastgen,cai2024pyramidkv,tang2024quest,xiao2024infllm,chen2023longlora,su2021roformer,chen2023positionalinterpolation,peng2023yarn}. SAFE-KV differs by studying whether a reused state is causally valid, not only whether it is compact or position-adjusted.

\paragraph{Retrieval, RAG, and robustness.}
Retrieval models and RAG benchmarks motivate the document-level setting \citep{lewis2021retrievalaugmentedgenerationknowledgeintensivenlp,karpukhin2020densepassage,guu2020realm,borgeaud2022retro,izacard2020fid,izacard2022atlas,asai2023selfrag,shi2023replug,petroni2021kilt,nakano2021webgpt,gao2023retrievalaugmented,gao2023alce,thakur2021beir,khattab2020colbert,santhanam2021colbertv2,nogueira2019passagereranking,izacard2021contriever,gao2022hyde}. Robustness work motivates the noise, negative-rejection, and position-sensitive dimensions \citep{chen2023rgb,yoran2023robustretrievalaugmented,shi2023distracted,cuconasu2024powernoise,liu2023lostmiddle}.

\paragraph{Reasoning, benchmarks, and internal state.}
Multi-hop and long-form QA benchmarks provide natural settings where evidence order and cross-document integration matter \citep{yang2018hotpotqadatasetdiverseexplainable,ho2020constructingmultihopqadataset,trivedi2022musique,kwiatkowski2019naturalquestions,joshi2017triviaqa,rajpurkar2016squad,thorne2018fever,bai2023longbench,hsieh2024ruler,dasigi2021qasper,kocisky2017narrativeqa,lin2021truthfulqa}. Work on reasoning, instruction following, and factual state supports the broader view that behavior depends on context-conditioned internal representations \citep{vaswani2017attention,devlin2019bert,brown2020gpt3,ouyang2022instructgpt,wei2022cot,wang2022selfconsistency,yao2022react,wang2022selfinstruct,geva2020ffnkv,meng2022rome,meng2022memit}. Model and tooling references document the open-model and implementation environment \citep{qwen2025qwen25technicalreport,jiang2023mistral,allal2025smollm2smolgoesbig,wolf2020huggingfacestransformersstateoftheartnatural,paszke2019pytorch,hoffmann2022chinchilla}.

\section{Related Work Comparison}
\label{sec:app-related-comparison}

Table~\ref{tab:app-related-comparison} summarizes how SAFE-KV relates to adjacent cache-reuse work. The comparison is coarse, since many systems support several deployment modes and SAFE-KV complements rather than replaces them; it highlights the evaluation target that distinguishes SAFE-KV.

\begin{table*}[t]
\centering
\footnotesize
\begin{tabular}{@{}>{\raggedright\arraybackslash}p{0.11\textwidth}>{\raggedright\arraybackslash}p{0.16\textwidth}>{\raggedright\arraybackslash}p{0.16\textwidth}>{\raggedright\arraybackslash}p{0.14\textwidth}>{\raggedright\arraybackslash}p{0.18\textwidth}>{\raggedright\arraybackslash}p{0.11\textwidth}@{}}
\toprule
Work & Reuse setting & Main goal & Full-prefill safety oracle? & Repair granularity? & Matched random controls? \\
\midrule
vLLM/Paged\-Attention & serving cache management & throughput/memory efficiency & no & no & no \\
LMCache & cache storage/transfer/reuse layer & system reuse & no & no & no \\
Prompt Cache & modular prompt reuse & latency reduction & limited & no & no \\
CacheBlend & non-prefix chunk fusion & serving efficiency + quality & partially, systems-focused & selective token recompute & not this framing \\
Cache-Craft & chunk-cache management for RAG & TTFT/throughput with quality control & no (task accuracy) & selective token recompute & no \\
KVLink & independent document cache reuse & speed + QA with adjustment/training & limited & no & no \\
SAFE-KV & non-prefix RAG document composition & behavior-preservation analysis & yes & document / window / suffix / token-local units compared & yes \\
\bottomrule
\end{tabular}
\caption{Positioning against adjacent cache-reuse work. SAFE-KV's contribution is the evaluation framing: full-prefill behavior preservation on the exact prompt, comparison of repair units, answer-level drift, and matched random controls. The CacheBlend and Cache-Craft rows summarize their published descriptions.}
\label{tab:app-related-comparison}
\end{table*}

\section{Artifact, Reproduction, and Computational Budget}
\label{sec:app-artifact}

\begin{table*}[t]
\centering
\footnotesize
\setlength{\tabcolsep}{4pt}
\begin{adjustbox}{max width=\textwidth}
\begin{tabular}{lrrrrr}
\toprule
$N$ & $p=0$ & $p=0.25$ & $p=0.5$ & $p=0.75$ & $p=1$ \\
\midrule
\multicolumn{6}{l}{\emph{No repair}} \\
1 & 1.000 [1.00, 1.00] @0.000 & 1.000 [1.00, 1.00] @0.000 & 1.000 [1.00, 1.00] @0.000 & 1.000 [1.00, 1.00] @0.000 & 1.000 [1.00, 1.00] @0.000 \\
2 & 1.000 [1.00, 1.00] @0.000 & 1.000 [1.00, 1.00] @0.000 & 1.000 [1.00, 1.00] @0.000 & 1.000 [1.00, 1.00] @0.000 & 1.000 [1.00, 1.00] @0.000 \\
4 & 1.000 [1.00, 1.00] @0.000 & 1.000 [1.00, 1.00] @0.000 & 1.000 [1.00, 1.00] @0.000 & 1.000 [1.00, 1.00] @0.000 & 1.000 [1.00, 1.00] @0.000 \\
8 & 1.000 [1.00, 1.00] @0.000 & 0.993 [0.98, 1.00] @0.000 & 0.993 [0.98, 1.00] @0.000 & 0.993 [0.98, 1.00] @0.000 & 1.000 [1.00, 1.00] @0.000 \\
16 & 1.000 [1.00, 1.00] @0.000 & 0.920 [0.85, 0.99] @0.000 & 0.927 [0.86, 0.99] @0.000 & 0.940 [0.89, 0.99] @0.000 & 0.800 [0.69, 0.91] @0.000 \\
\midrule
\multicolumn{6}{l}{\emph{Query doc-only K1}} \\
1 & 1.000 [1.00, 1.00] @0.000 & 1.000 [1.00, 1.00] @0.000 & 1.000 [1.00, 1.00] @0.000 & 1.000 [1.00, 1.00] @0.000 & 1.000 [1.00, 1.00] @0.000 \\
2 & 1.000 [1.00, 1.00] @0.405 & 1.000 [1.00, 1.00] @0.405 & 1.000 [1.00, 1.00] @0.405 & 1.000 [1.00, 1.00] @0.405 & 1.000 [1.00, 1.00] @0.405 \\
4 & 1.000 [1.00, 1.00] @0.223 & 1.000 [1.00, 1.00] @0.223 & 1.000 [1.00, 1.00] @0.223 & 1.000 [1.00, 1.00] @0.223 & 1.000 [1.00, 1.00] @0.223 \\
8 & 1.000 [1.00, 1.00] @0.116 & 1.000 [1.00, 1.00] @0.116 & 0.993 [0.98, 1.00] @0.116 & 0.993 [0.98, 1.00] @0.116 & 1.000 [1.00, 1.00] @0.116 \\
16 & 1.000 [1.00, 1.00] @0.055 & 0.920 [0.85, 0.99] @0.055 & 0.927 [0.86, 0.99] @0.055 & 0.960 [0.92, 1.00] @0.055 & 0.840 [0.74, 0.94] @0.055 \\
\midrule
\multicolumn{6}{l}{\emph{Query window w4 K1}} \\
1 & 1.000 [1.00, 1.00] @0.000 & 1.000 [1.00, 1.00] @0.000 & 1.000 [1.00, 1.00] @0.000 & 1.000 [1.00, 1.00] @0.000 & 1.000 [1.00, 1.00] @0.000 \\
2 & 1.000 [1.00, 1.00] @0.405 & 1.000 [1.00, 1.00] @0.405 & 1.000 [1.00, 1.00] @0.405 & 1.000 [1.00, 1.00] @0.405 & 1.000 [1.00, 1.00] @0.405 \\
4 & 1.000 [1.00, 1.00] @0.639 & 1.000 [1.00, 1.00] @0.639 & 1.000 [1.00, 1.00] @0.639 & 1.000 [1.00, 1.00] @0.639 & 1.000 [1.00, 1.00] @0.639 \\
8 & 1.000 [1.00, 1.00] @0.454 & 1.000 [1.00, 1.00] @0.454 & 1.000 [1.00, 1.00] @0.454 & 1.000 [1.00, 1.00] @0.454 & 1.000 [1.00, 1.00] @0.454 \\
16 & 1.000 [1.00, 1.00] @0.240 & 0.913 [0.85, 0.98] @0.240 & 0.933 [0.87, 1.00] @0.240 & 0.953 [0.90, 1.00] @0.240 & 0.960 [0.91, 1.00] @0.240 \\
\midrule
\multicolumn{6}{l}{\emph{Query suffix K1}} \\
1 & 1.000 [1.00, 1.00] @0.000 & 1.000 [1.00, 1.00] @0.000 & 1.000 [1.00, 1.00] @0.000 & 1.000 [1.00, 1.00] @0.000 & 1.000 [1.00, 1.00] @0.000 \\
2 & 1.000 [1.00, 1.00] @0.405 & 1.000 [1.00, 1.00] @0.405 & 1.000 [1.00, 1.00] @0.405 & 1.000 [1.00, 1.00] @0.405 & 1.000 [1.00, 1.00] @0.405 \\
4 & 1.000 [1.00, 1.00] @0.639 & 1.000 [1.00, 1.00] @0.639 & 1.000 [1.00, 1.00] @0.639 & 1.000 [1.00, 1.00] @0.639 & 1.000 [1.00, 1.00] @0.639 \\
8 & 1.000 [1.00, 1.00] @0.785 & 1.000 [1.00, 1.00] @0.785 & 1.000 [1.00, 1.00] @0.785 & 1.000 [1.00, 1.00] @0.785 & 1.000 [1.00, 1.00] @0.785 \\
16 & 1.000 [1.00, 1.00] @0.898 & 1.000 [1.00, 1.00] @0.898 & 1.000 [1.00, 1.00] @0.898 & 1.000 [1.00, 1.00] @0.898 & 1.000 [1.00, 1.00] @0.898 \\
\midrule
\multicolumn{6}{l}{\emph{Risk suffix K1}} \\
1 & 1.000 [1.00, 1.00] @0.000 & 1.000 [1.00, 1.00] @0.000 & 1.000 [1.00, 1.00] @0.000 & 1.000 [1.00, 1.00] @0.000 & 1.000 [1.00, 1.00] @0.000 \\
2 & 1.000 [1.00, 1.00] @0.405 & 1.000 [1.00, 1.00] @0.405 & 1.000 [1.00, 1.00] @0.405 & 1.000 [1.00, 1.00] @0.405 & 1.000 [1.00, 1.00] @0.405 \\
4 & 1.000 [1.00, 1.00] @0.629 & 1.000 [1.00, 1.00] @0.629 & 1.000 [1.00, 1.00] @0.629 & 1.000 [1.00, 1.00] @0.629 & 1.000 [1.00, 1.00] @0.629 \\
8 & 1.000 [1.00, 1.00] @0.753 & 1.000 [1.00, 1.00] @0.753 & 0.993 [0.98, 1.00] @0.753 & 0.993 [0.98, 1.00] @0.753 & 1.000 [1.00, 1.00] @0.753 \\
16 & 1.000 [1.00, 1.00] @0.686 & 0.967 [0.92, 1.00] @0.686 & 0.980 [0.95, 1.00] @0.686 & 0.987 [0.96, 1.00] @0.686 & 0.980 [0.94, 1.00] @0.686 \\
\midrule
\multicolumn{6}{l}{\emph{Risk suffix K2}} \\
1 & 1.000 [1.00, 1.00] @0.000 & 1.000 [1.00, 1.00] @0.000 & 1.000 [1.00, 1.00] @0.000 & 1.000 [1.00, 1.00] @0.000 & 1.000 [1.00, 1.00] @0.000 \\
2 & 1.000 [1.00, 1.00] @0.405 & 1.000 [1.00, 1.00] @0.405 & 1.000 [1.00, 1.00] @0.405 & 1.000 [1.00, 1.00] @0.405 & 1.000 [1.00, 1.00] @0.405 \\
4 & 1.000 [1.00, 1.00] @0.636 & 1.000 [1.00, 1.00] @0.636 & 1.000 [1.00, 1.00] @0.636 & 1.000 [1.00, 1.00] @0.636 & 1.000 [1.00, 1.00] @0.636 \\
8 & 1.000 [1.00, 1.00] @0.778 & 1.000 [1.00, 1.00] @0.778 & 1.000 [1.00, 1.00] @0.778 & 1.000 [1.00, 1.00] @0.778 & 1.000 [1.00, 1.00] @0.778 \\
16 & 1.000 [1.00, 1.00] @0.796 & 0.980 [0.94, 1.00] @0.796 & 0.993 [0.98, 1.00] @0.796 & 1.000 [1.00, 1.00] @0.796 & 1.000 [1.00, 1.00] @0.796 \\
\midrule
\multicolumn{6}{l}{\emph{Mask-informed earliest-stale oracle}} \\
1 & 1.000 [1.00, 1.00] @0.000 & 1.000 [1.00, 1.00] @0.000 & 1.000 [1.00, 1.00] @0.000 & 1.000 [1.00, 1.00] @0.000 & 1.000 [1.00, 1.00] @0.000 \\
2 & 1.000 [1.00, 1.00] @0.000 & 1.000 [1.00, 1.00] @0.102 & 1.000 [1.00, 1.00] @0.200 & 1.000 [1.00, 1.00] @0.302 & 1.000 [1.00, 1.00] @0.405 \\
4 & 1.000 [1.00, 1.00] @0.000 & 1.000 [1.00, 1.00] @0.254 & 1.000 [1.00, 1.00] @0.456 & 1.000 [1.00, 1.00] @0.564 & 1.000 [1.00, 1.00] @0.655 \\
8 & 1.000 [1.00, 1.00] @0.000 & 1.000 [1.00, 1.00] @0.490 & 1.000 [1.00, 1.00] @0.681 & 1.000 [1.00, 1.00] @0.765 & 1.000 [1.00, 1.00] @0.814 \\
16 & 1.000 [1.00, 1.00] @0.000 & 1.000 [1.00, 1.00] @0.715 & 1.000 [1.00, 1.00] @0.835 & 1.000 [1.00, 1.00] @0.872 & 1.000 [1.00, 1.00] @0.904 \\
\bottomrule
\end{tabular}
\end{adjustbox}
\caption{Full staleness-probability $\times$ document-count grid (26{,}250 rows; validator passed with zero invariant failures). Each cell reports normalized answer preservation relative to the all-fresh reference, its example-clustered 95\% interval, and the mean recomputed context-token fraction (@). Fifty paired examples per $N$, mask seeds 13/29/47. The mask-informed oracle recomputes from the earliest truly mismatched slice and is a sanity check of the intervention, not a deployable policy. Actual forwarding; Qwen2.5-7B-Instruct; one NVIDIA L40S.}
\label{tab:app-stale-full}
\end{table*}

{\raggedright %
\paragraph{Package layout.}
The artifact is the version-locked repository \url{https://github.com/XingjianZhang-dev/safe-kv} (tag \texttt{v1.0.0}), which contains every run reported in the paper. Its root holds: \texttt{code/} (the \texttt{safekv} package, scripts, and tests), \texttt{configs/}, \texttt{data/} (processed HotpotQA-500/2Wiki-500 JSONL, synthetic diagnostics, position sweeps), \texttt{results/} (main donor-copy runs), \texttt{results\_rebuttal/} (actual-forwarding, reconciliation, staleness, and latency runs with per-example outputs, manifests, and validator reports, plus their \texttt{data/} variants), \texttt{results\_august/} (the query-attention baseline runs, their datasets, and the \texttt{safekv\_ipm} code that produced them), \texttt{paper\_reference/} (table sources and figure data), the README, reproduction guide, manifests, checksums (\texttt{SHA256SUMS}), result listing, test log, \texttt{LICENSE}, and \texttt{NOTICE}. Raw per-example dumps of four run families (232\,MB) are attached to release \texttt{v1.0.0} as a tarball that \path{scripts/fetch_raw_outputs.sh} restores in place; every summary the paper uses is in the repository itself.

\paragraph{Key configs and scripts.}
The main Qwen2.5-7B external configs are \path{configs/acl_qwen25_7b_hotpotqa_500.yaml} and \path{configs/acl_qwen25_7b_2wiki_500.yaml}; their legacy \texttt{risk} block is not read by the selector (Appendix~\ref{sec:app-selector}). Answer-level runs use \path{code/scripts/run_answer_eval.py} (methods \texttt{actual\_recompute::*} for actual forwarding); score-neutral and nested inputs are built by \path{code/scripts/build_rebuttal_variants.py}; the staleness grid by \path{code/scripts/run_staleness_doccount_sweep.py}; latency by \path{code/scripts/run_rebuttal_latency_trace.py} with \path{code/scripts/validate_rebuttal_latency_results.py} and \path{code/scripts/summarize_latency_paired_bootstrap.py}; bootstrap intervals by \path{code/scripts/bootstrap_answer_eval_ci.py}; held-out calibration by \path{code/scripts/heldout_method_calibration.py}; the token-local baseline by \path{results_august/code/scripts/run_query_attention_local_baseline.py}; the oracle KV-drift diagnostic by \path{code/scripts/run_oracle_kv_drift_local_baseline.py}. \texttt{FULL\_REPRODUCTION.md} gives the exact command for every table.

\paragraph{Key result directories.}
Main runs: \path{results/acl_qwen25_7b_hotpotqa_suffix_main_500}, \path{results/acl_qwen25_7b_2wiki_suffix_main_500} (with \path{bootstrap_ci/} and \path{heldout_method_calibration_safekv_only/}), \path{results/acl_qwen25_7b_2wiki_downstream_window_100}, \path{results/acl_qwen25_7b_position_noise16_50x5_suffix}, the \texttt{noise\_sweep}/\texttt{noise16} directories, \path{results/acl_mistral7b_noisy_suffix_250_balanced}. Actual forwarding, staleness, and latency: \path{results_rebuttal/actual_recompute_{hotpot,2wiki}_500}, \path{results_rebuttal/actual_vs_copy_{hotpot,2wiki}_500_l40s_v2}, \path{results_rebuttal/neutral_selector_actual_*_shuffle{101,202,303}_100} with \path{neutral_selector_actual_3seed_validation}, \path{results_rebuttal/staleness_doccount_full_50_v3}, the five \path{results_rebuttal/latency_*_50_v1} directories, \path{results_rebuttal/quality_noise64_actual_riskk1_50_v1}, \path{results_rebuttal/oracle_kv_drift_2wiki_neutral_shuffle101_100_v3}, \path{results_rebuttal/cpu_selector_audit_l40s_host}. Token-local baseline: \path{results_august/run_registry/qwen25_query_attention_hotpotqa_{natural,shuffle,bm25}_test_n100_v1}. Appendix~\ref{sec:app-result-inventory} maps every result family to its use.

\par}

\paragraph{Tests.}
In a fresh clone with the pinned CPU dependencies, \texttt{python -m pytest -q code/tests} reports \cpuTestsPassed{} passed and \cpuTestsSkipped{} skipped (one environment-dependent test); the log is \texttt{TEST\_LOG.txt}.

\paragraph{Computational budget (estimate).}
The budget is estimated from run manifests and directory timestamps. The main donor-copy experiments (HotpotQA-500 and 2Wiki-500 external runs, controlled diagnostics, position sweeps, cross-model checks, prototype latency, and development reruns) used on the order of 100--160 single-GPU hours on an L40S-class GPU, with about 120 as the point estimate. The actual-forwarding experiments (actual-forwarding reruns, six score-neutral runs, the 26{,}250-row staleness grid, all-document controls, five latency workloads, the Noise64 quality pairing, and the oracle diagnostic; about 41{,}000 answer-level rows and 7{,}500 timed calls in total) used one NVIDIA L40S for an estimated 10--20 further GPU-hours. The three localized-baseline runs used one NVIDIA RTX PRO 6000 Blackwell GPU for 2.8 hours each (run-manifest wall times of 10{,}209, 10{,}245, and 10{,}291 seconds). All compute was inference and evaluation; no model was trained.

\paragraph{Scope of claims.}
The paper's claims are 7B-scale behavior-preservation results under two labelled protocols, with paired bootstrap support, held-out calibration, a staleness-by-document-count analysis, tested localized baselines, and batch-1 latency measurements. Production throughput, QA accuracy, selector reliability on arbitrary retrievers, and head-to-head comparisons with Cache-Craft or CacheBlend are outside this scope.

\paragraph{Availability and licensing.}
The artifact is available at \url{https://github.com/XingjianZhang-dev/safe-kv}. The code, configs, and derived result files are released under the MIT License (\texttt{LICENSE}). Model weights (Qwen2.5-7B-Instruct, Mistral-7B-Instruct-v0.3, and the development models) are not redistributed; the public benchmark data are not redistributed beyond the processed HotpotQA and 2WikiMultihopQA examples used in the experiments, which stay under the original datasets' terms (HotpotQA: CC BY-SA 4.0); third-party dependencies are under their own licenses (\texttt{NOTICE}).

\section{Staleness Probability $\times$ Document Count: Full Grid}
\label{sec:app-stale-full}

Table~\ref{tab:app-stale-full} lists every cell of the factorial behind Table~\ref{tab:staleness}: seven methods, five document counts, five mismatch probabilities, 150 rows per cell (50 paired examples $\times$ three mask seeds), with example-clustered 95\% intervals and the mean recomputed context fraction. The all-fresh chunk-matched reference agrees with one-shot full prefill in every cell, and the mask-informed earliest-stale oracle recovers every cell, which confirms the intervention itself (it reads the sampled mask and serves only as a sanity check). Paired deltas against no repair at $N{=}16$, $p{=}1$ (normalized answer preservation, example bootstrap): query window w4 $+0.16$ [0.06, 0.26], risk K1 $+0.18$ [0.06, 0.30], risk K2 $+0.20$ [0.10, 0.32], query suffix K1 $+0.20$ [0.10, 0.32], oracle $+0.20$ [0.10, 0.32]. At $N{=}16$, $p{=}0.25$ the query window w4 delta is $-0.007$ [$-0.027$, $0.013$]: at low mismatch the window adds nothing over reuse. Because retrieval scores are zero in these inputs and the nested documents tie on the lexical features, the query selector starts at the first document in 42, 45, and 43 of 50 examples for $N=4$, 8, and 16, and risk K1 in 42, 40, and 19; these rows therefore mostly compare early-start windows and suffixes (Appendix~\ref{sec:app-selector}).

\end{document}